\documentclass[a4paper,fleqn]{cas-dc}

\usepackage[numbers]{natbib}

\usepackage{float}                  % 图片浮动位置
\usepackage{subfig} 
\usepackage[mathscr]{euscript} %花体公式
\usepackage{amsmath}
\usepackage{makecell}
\usepackage[switch]{lineno}
\usepackage{color}
\usepackage{multirow}\usepackage{makecell}

\def\tsc#1{\csdef{#1}{\textsc{\lowercase{#1}}\xspace}}
\tsc{WGM}
\tsc{QE}
\begin{document}
\let\WriteBookmarks\relax
\def\floatpagepagefraction{1}
\def\textpagefraction{.001}

\shorttitle{}    

\shortauthors{}  

\title [mode = title]{Wildfire Smoke Detection System: Model Architecture, Training Mechanism, and Dataset}


\author[1]{Chong Wang}[orcid=0009-0008-5190-7792]
\ead{wchong@ustc.edu.cn}
\credit{Conceptualization, Formal Analysis, Investigation, Methodology, Software, Validation, Visualization, Writing - Original Draft}

\author[2]{Chen Xu}[orcid=0000-0003-4163-6024]
\ead{xuchenustc@mail.ustc.edu.cn}
\credit{Data Curation, Validation, Visualization, Writing - Review \& Editing}

\author[1,2]{Adeel Akram}[orcid=0000-0001-9901-0716]
\ead{dradeel@ustc.edu.cn}
\credit{Validation, Writing - Review \& Editing}

\author[1]{Zhong Wang}[orcid=0000-0001-7987-4883]
\ead{zhongw@ustc.edu.cn}
\credit{Project administration}

\author[2,3]{Zhilin Shan}[orcid=0000-0002-7236-3981]
\ead{shanuqn@163.com}
\credit{Resources}

\author[1]{Qixing Zhang}[orcid=0000-0002-8784-8674]
\cormark[1]
\ead{qixing@ustc.edu.cn}
\credit{Conceptualization, Funding Acquisition, Resources, Supervision}
\cortext[1]{Corresponding author}



\affiliation[1]{organization={State Key Laboratory of Fire Science, University of Science and Technology of China},
            addressline={}, 
            city={Hefei},
%          citysep={}, % Uncomment if no comma needed between city and postcode
            postcode={230026}, 
            state={Anhui},
            country={China}}
\affiliation[2]{organization={Institute of Advanced Technology, University of Science and Technology of China},
            addressline={}, 
            city={Hefei},
%          citysep={}, % Uncomment if no comma needed between city and postcode
            postcode={230031}, 
            state={Anhui},
            country={China}}
\affiliation[3]{organization={iFIRE TEK Co., Ltd.},
            addressline={}, 
            city={Hefei},
%          citysep={}, % Uncomment if no comma needed between city and postcode
            postcode={230031}, 
            state={Anhui},
            country={China}}

\begin{abstract}
Vanilla Transformers focus on semantic relevance between mid- to high-level features and are not good at extracting smoke features as they overlook subtle changes in low-level features like color, transparency, and texture which are essential for smoke recognition. To address this, we propose the Cross Contrast Patch Embedding (CCPE) module based on the Swin Transformer. This module leverages multi-scale spatial contrast information in both vertical and horizontal directions to enhance the network's discrimination of underlying details. By combining Cross Contrast with Transformer, we exploit the advantages of Transformer in global receptive field and context modeling while compensating for its inability to capture very low-level details, resulting in a more powerful backbone network tailored for smoke recognition tasks. Additionally, we introduce the Separable Negative Sampling Mechanism (SNSM) to address supervision signal confusion during training and release the SKLFS-WildFire Test dataset, the largest real-world wildfire testset to date, for systematic evaluation. Extensive testing and evaluation on the benchmark dataset FIgLib and the SKLFS-WildFire Test dataset show significant performance improvements of the proposed method over baseline detection models. The code and data are available at \url{github.com/WCUSTC/CCPE}. 
\end{abstract}


%\item Cross-contrast patch embedding enhances the detail modeling of Transformers.
%\item Separable negative sampling alleviates boundary-ambiguity-induced challenges.
%\item A larger-scale wildfire smoke dataset facilitates future research.

% Keywords
\begin{keywords}
WildFire Detection \sep Smoke Detection \sep Cross Contrast \sep Patch Embedding \sep Separable Negative Sampling \sep Smoke Transformer
\end{keywords}

\maketitle

\section{Introduction}
\label{}
Visual wildfire detection refers to the utilization of imaging devices to capture image or video signals, followed by the application of intelligent algorithms to find out the potential wildfire cues. The two critical indicators for determining the presence of wildfires are smoke and flames. Smoke often manifests earlier in the wildfire occurrence process than flames and has the advantage of being less susceptible to vegetation obstruction, making it increasingly intriguing for researchers~\cite{chaturvedi2022survey}. In terms of imaging methods, wildfire detection typically involves using visible light cameras or infrared cameras to capture images or videos. Visible light-based wildfire detection is favored for its cost-effectiveness and broad coverage, gaining popularity among researchers~\cite{yar2023effective,xue2022small}. However, due to the complex background interferences in open scenes, visible light-based wildfire detection demands higher requirements for intelligent algorithms, necessitating further research efforts to advance this technology. 

%Wildfire detection can be classified into three application scenarios based on the camera mounting platform: ground-mounted~\cite{zhang2018wildland,li2020image}, UAV-mounted~\cite{chen2022wildland,jiao2019deep}, and satellite-mounted~\cite{rostami2022active,rashkovetsky2021wildfire,kang2022deep}. This study primarily discusses smoke detection based on visible light images. Figure~\ref{fig:1} shows examples of visible-light smoke images captured from different platforms. Although the dataset is collected using ground-mounted cameras, the findings remain inspirational for wildfire detection based on satellite remote sensing or unmanned aerial vehicle (UAV) imaging.
%\begin{figure}[t]
%	\centering
%	\includegraphics[width=0.49\textwidth]{1.pdf}
%	\caption{Examples of wildfire smoke images captured from different platforms. (a-c) are images of smoke captured by satellites, UAVs, and ground-based cameras, respectively.}	
%	\label{fig:1}
%\end{figure}

Over the past decade, researchers have developed a large number of deep learning-based wildfire recognition, detection and even segmentation models. For example, ~\cite{muhammad2018efficient, gu2019deep} have proposed fire image classification networks based on deep convolutional neural networks, while~\cite{xue2022small,lin2023semi} added classical detector heads on the basis of CNNs backbone to identify or locate smoke or flame. In recent years, Transformer-based models have demonstrated comparable or superior performance to CNNs on many tasks. Therefore, some researchers have tried to apply Transformers to the field of fire detection. For example, Khudayberdiev~\emph{et.al.} \cite{khudayberdiev2022fire} proposed to use Swin Transformer~\cite{liu2021swin} as the backbone network to realize the classification of fire images. In \cite{hong2024wildfire}, a variety of classical backbone networks including ResNet~\cite{he2016deep}, MobileNet~\cite{howard2017mobilenets}, Swin Transformer~\cite{liu2021swin} and ConvNeXt~\cite{liu2022convnet} are used to realize wildfire detection with self-designed detection heads, and it is proven in experiments that the Transformer model has no obvious advantage over CNNs. We have observed similar phenomena in our experiments. Why do Transformer-based models work well in other tasks but fail in wildfire detection?

Through the analysis of a large amount of real fire data, we found that smoke, one of the most typical early cues of fire, has special properties that are different from entity objects. An important basis for judging the presence of fire smoke is the spatial distribution of transparency, color and texture. These features are generally extracted at the bottom of the deep neural networks. The Transformer network establishes the correlation between different areas through the attention mechanism, and has unique advantages in modeling long-distance dependencies and contextual correlation, but it has a poor ability to capture low-level details. Based on this observation, this article proposes the Cross Contrast Patch Embedding (CCPE) module to promote Swin Transformer's ability to distinguish the underlying smoke texture. Specifically, we sequentially cascade a vertical multi-spatial frequency contrast structure and a horizontal multi-spatial frequency contrast structure within the Patch Embedding, and use the cascaded spatial contrast results to enhance the original embedding results and input them into the subsequent network. We found that this simple design can bring extremely significant performance improvements with an almost negligible increase in computational effort.

\begin{figure*}[t]
	\centering
	\includegraphics[width=0.99\textwidth]{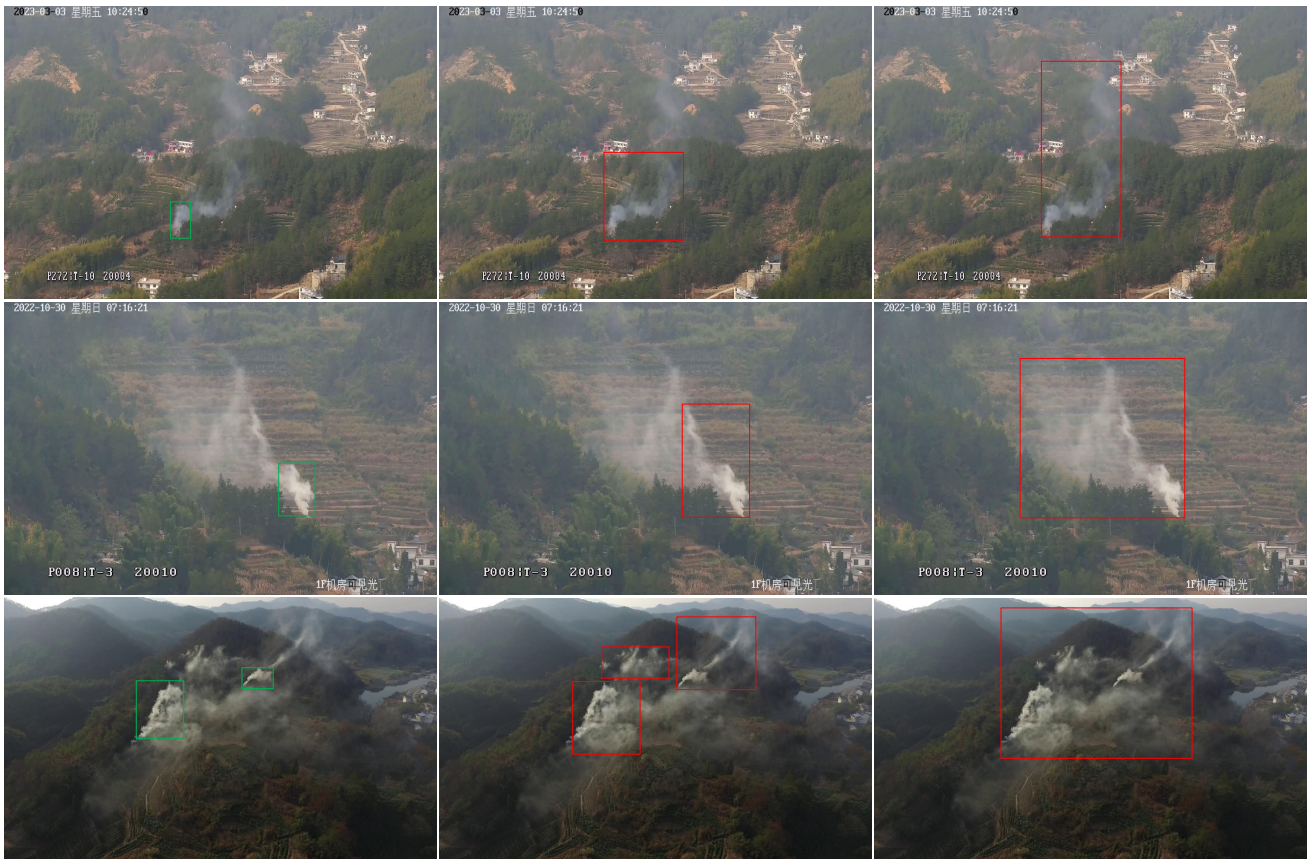}
	\caption{Display of fuzzy characteristics of smoke boundary and uncertainty of annotation. The \textcolor{green}{green} bounding boxes are the manual labeled Ground Truth, while the \textcolor{red}{red} are the possible candidate boxes. Following the general object detection paradigm, the red box is likely to be assigned a negative label, causing difficulties in model training.}	
	\label{fig:introduction_1}
\end{figure*}

The main difference between wildfire smoke detection and general object detection tasks is the ambiguity of smoke object boundaries. On the one hand, wildfire smoke detection in open scenes often encounters false alarms and requires the addition of a large number of negative error-prone image samples, that is, the images do not contain wildfire but there are objects with high appearance similarity to smoke. Since the number of negative image samples far exceeds the number of wildfire images, and error-prone objects only account for a small proportion of the image area. The natural idea is to use Online Hard Example Mining (OHEM)~\cite{shrivastava2016training} to focus on confusing areas when sampling negative proposals, thereby improving the accuracy of the detection model. On the other hand, smoke objects show different transparency at different spatial locations due to different concentrations. It is difficult to clearly define the density boundary between the foreground and background of smoke during manual annotation. This leads to ambiguity in the labeling range of the smoke bounding boxes. In the classic object detection framework, the label assignment of proposals needs to be determined based on the ground-truth boxes. As shown in Figure~\ref{fig:introduction_1}, the green box is the manually labeled smoke foreground, and the red boxes represent the controversial proposals, which are practically exhaustive and cannot be annotated one by one. During the model training phase, it is inappropriate to assign background labels to proposals represented by red boxes. When the OHEM strategy encounters ambiguous smoke objects, most of the negative proposals acquired during training will be ambiguous. To solve this problem, this paper proposes a Separable Negative Sampling Mechanism (SNSM). Specifically, the positive and negative images in the batch are separated during training, and a small number of negative proposals are collected from the positive images with wildfire smoke, and OHEM is used to collect the confusing areas in the negative images without wildfire smoke. Separable negative instance sampling can increase recall and improve model performance in high recall intervals.

Open-scene wildfire detection lacks a large-scale test dataset for comparison and validation. For instance, the widely employed public dataset, the Fire Detection Dataset \cite{foggia2015real},comprises 149 videos, of which 74 videos feature fire smoke. The recently released FIgLib dataset \cite{dewangan2022figlib} encompasses 24.8K images from 315 videos (including omitted images), while the test set only comprises 4,880 images sourced from 45 videos. And, some existing researches \cite{hu2018real,lin2023semi} only report results on undisclosed datasets. The insufficient scale of the public datasets leads to large fluctuations in test results, and the credibility of the experimental results is questionable. This article discloses a large-scale test set, named SKLFS-WildFire Test, which contains 3,309 short video clips, including 340 real wildfire videos, and the rest are negative examples of no fire incidents. We obtained a total of 50,735 images by sampling frames with intervals, of which 3,588 were images with smoke. False positives are the most critical issue in wildfire detection that affects the user experience, so our test set contains a large number of negative sample images with a large number of interference objects that resemble the appearance of the smoke. It provides a benchmark for testing the performance of models in unpredictable environments.

To sum up, the \textbf{main contributions} of this paper are twofold:
\begin{enumerate}

\item We propose a new Transformer-based wildfire smoke detector. The issue of the insufficient ability of the Transformer backbone to capture smoke details is addressed through the Cross Contrast Patch Embedding module. Additionally, the challenge of difficult label assignment caused by the unclear boundaries of smoke objects is mitigated by the Separable Negative Sampling Mechanism.
\item SKLFS-WildFire Test, the largest real wildfire test dataset to date, was released, surpassing even most of the published wildfire training sets.
\end{enumerate} 
Our algorithm underwent extensive evaluation on 1200 ground-mounted cameras, affirming its efficacy and reliability over prolonged periods.

\section{Related Work}
\begin{table*}
\begin{center}
\caption{Overview of publicly available fire datasets and the SKLFS-WildFire Test.}
\renewcommand\arraystretch{2}
\resizebox{0.99\textwidth}{!}{
\begin{tabular}{l|c|c|c|c|c}
\hline
Datasets       &  \quad Data Format \quad     & \quad   Flame/Smoke \quad   &  \quad   Positive Number \quad  &\quad  Total Number\quad  &\quad  Annotation Form\quad \\
\hline
\makecell[l]{Fire Detection Dataset \cite{foggia2015real}}  &   Video      &    Flame    &     14 Videos   &  31 Videos  &  Video Level \\
\makecell[l]{Smoke Detection Dataset \cite{foggia2015real}}  &   Video      &    Smoke    &     74 Videos   &  149 Videos  &  Video Level \\
\makecell[l]{Bowfire \cite{chino2015bowfire}} &   Image      &    Flame    &     119 Images   &  226 Images  &  Pixel Level \\
\makecell[l]{Forest Fire Dataset \cite{khan2022deepfire}}  &   Image      &    No distinction     &     950 Images   &  1,900 Images  &  Image Level \\
\makecell[l]{FireFlame \cite{Fire-Flame-Dataset}} & Image  & Flame\&Smoke  &   2,000 Images   &  3,000 Images  &  Image Level \\
\makecell[l]{FireNet Dataset \cite{jadon2019firenet}} & Image\&Video  & No distinction &  46 Videos & \makecell[c]{62 Videos\&\\ 160 Neg. images} &Image Level \\
\makecell[l]{FIgLib Dataset \cite{dewangan2022figlib}} & Video  & Smoke  & 315 Videos & 315 Videos  & Image Level \\
\makecell[l]{SKLFS-WildFire Test Dataset\\ (OURS)} & Image & Flame\&Smoke &  3,588 Images & 50,735 Images  & Bbox Level \\
\hline
\end{tabular}
}
\label{tab:dataset}
\end{center}
\end{table*}

\subsection{Wildfire Detection Methods}
There have been a large number of public reports of deep learning-based wildfire detection methods, and remarkable progress has been made. However, the existing methods of wildfire detection are still not satisfactory in practical applications. The characteristics of wildfire detection missions, the challenges they face, and the obstacles to their practical application are still lacking in more in-depth and adequate discussion.
Most of the studies focus on the application of mainstream backbones, adjustments of parameters, and performance-efficiency trade-offs. For example, \cite{yin2017deep,namozov2018efficient,muhammad2018early} use CNN to extract deep features instead of manual features for image smoke detection, and \cite{yin2017deep} emphasizes the use of Batch Normalization(BN) in the network. Some reasearches \cite{zhang2018wildland, jiao2019deep, li2020image} use classical backbone networks or detectors such as MobileNetV2~\cite{sandler2018mobilenetv2}, YOLOV3~\cite{redmon2018yolov3}, SSD~\cite{liu2016ssd}, Faster R-CNN~\cite{ren2016faster} for smoke/flame recognition or detection. For model efficiency, \cite{muhammad2018convolutional, jadon2019firenet, muhammad2019efficient} have all designed new CNN backbone networks for fire recognition, and they emphasize that the newly designed network is highly efficient. 
Although the efficiency of the model is of great significance, the performance of the wildfire smoke detection model has not yet reached a satisfactory level. i.e. too many false positives or low recall. Xue \emph{et.al.} \cite{xue2022small} try to improve the model's small object detection capabilities by improving YOLOV5~\cite{jocher2022ultralytics}. However, in the general object detection field, there are numerous discussions and solutions regarding small object detection. While the detection of small smoke targets in the early stages of wildfires is challenging, it is not the primary issue in wildfire detection. Dewangan \emph{et al}.  \cite{dewangan2022figlib} proposed wildfire smoke detection methods utilizing spatio-temporal modeling and Bhamra \emph{et al}. \cite{bhamra2023multimodal} proposed to incorporate multimodal data such as weather and satellite data, achieving remarkable performance. However, the baseline approach relying on static images for smoke detection still faces unresolved issues. Recently, researchers have also tried to use transformers to improve the model expression ability of fire recognition tasks. Khudayberdiev \emph{et.al.} \cite{khudayberdiev2022fire} use Swin Transformer as the backbone network to classify fire images without making any improvements. Hong \emph{et.al.} \cite{hong2024wildfire} use various CNNs and transformer networks as backbones for fire recognition tasks, including Swin Transformer, DeiT \cite{touvron2021training}, ResNet, MobileNet, EfficientNet \cite{tan2019efficientnet}, ConvNeXt \cite{liu2022convnet}, etc., to compare the effects of different backbones. Experimental results show that the transformer backbones have no significant advantage over the CNN backbones. This is consistent with our observations.

\subsection{Wildfire Datasets}

\begin{figure*}[t]
	\centering
	\includegraphics[width=0.99\textwidth]{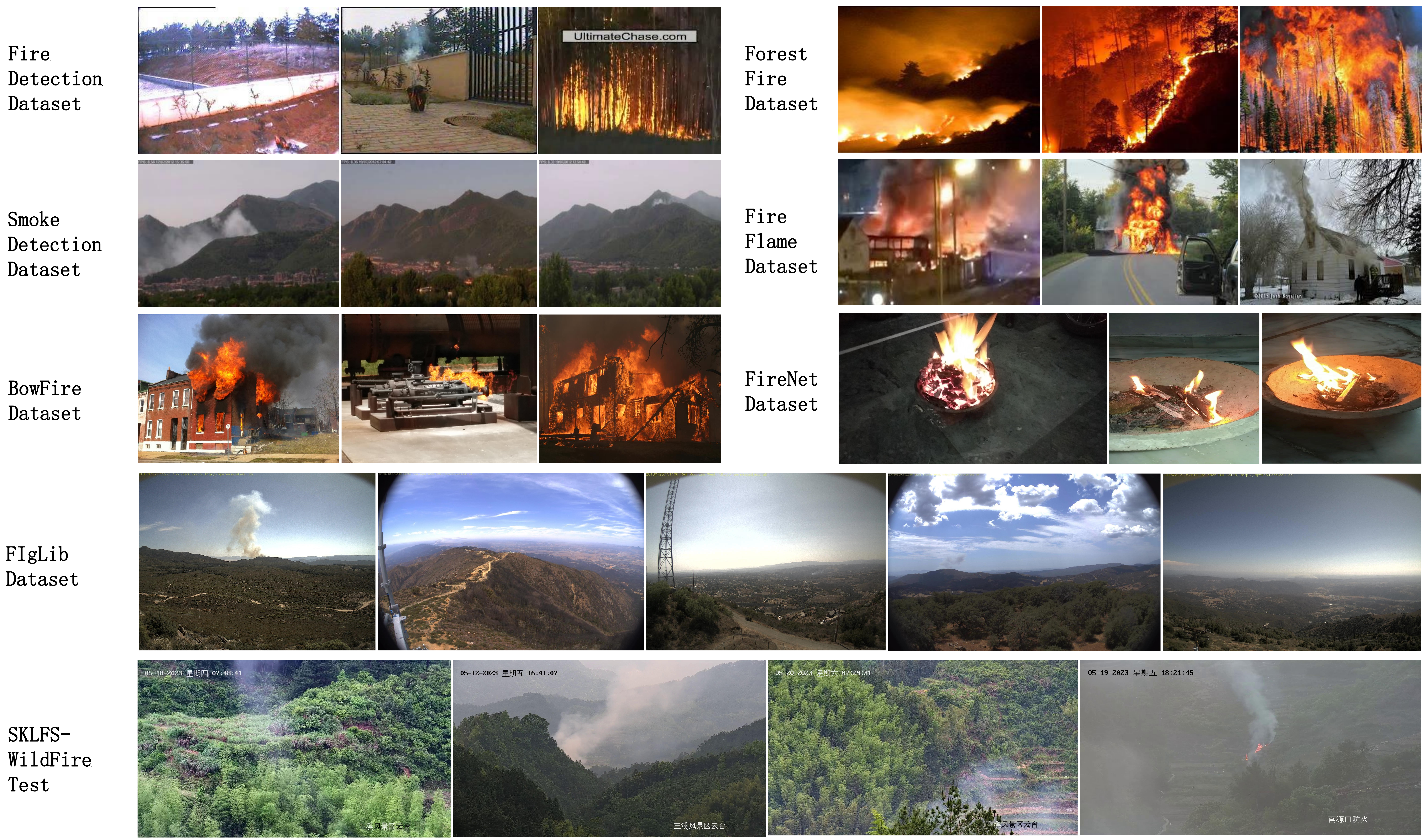}
	\caption{Illustration of samples from  publicly available datasets and the SKLFS-WildFire Test. Most of the existing data sets are not realistic enough in the scene and do not meet the actual needs of early fire warning. The SKLFS-WildFire Test offers significant advantages in both quality and scale.}	
	\label{fig:introduction_2}
\end{figure*}

%The research on wildfire detection is mainly based on self-built and non-public datasets.
Publicly available wildfire datasets are few and generally small in size. Table \ref{tab:dataset} summarizes the commonly used public datasets for wildfire recognition or detection. The Fire Detection Dataset \cite{foggia2015real} published by Pasquale \emph{et al.} is one of the most commonly used fire test sets, containing 31 video clips, 14 of which contain flames. It is worth mentioning that, videos with only smoke but no flame are considered as negative samples in this dataset. They also published the Smoke Detection Dataset, which contains 149 video clips, and 74 videos contain Smoke. However, to the best of our knowledge, no published papers have been found to report experimental results on the Smoke Detection Dataset. Chino \emph{et al.} \cite{chino2015bowfire} made a dataset publicly available, called Bowfire, which contains 226 images, of which 119 contain flames and 107 are negative samples without flames, and are equipped with pixel-level semantic segmentation labels. Ali \emph{et al.} \cite{khan2022deepfire} collected the Forest-Fire Dataset, which contains a total of 1900 images in the training set and the test set, of which 950 are images with fire. In \cite{Fire-Flame-Dataset}, a dataset named FireFlame is released, which contains three categories: Fire (namely Flame in this paper), smoke, and neutral, with 1000 images each, for a total of 3000 images. Arpit \emph{et al.} \cite{jadon2019firenet} collected a publicly available dataset, named FireNet, which contains a total of 108 video clips and 160 images that are prone to error detection. Dewangan \emph{et al}. \cite{dewangan2022figlib} introduced FIgLib, which to our knowledge is the largest wildfire recognition dataset, comprising 315 videos, of which 270 are usable. Building upon FIgLib, Bhamra \emph{et al}. \cite{bhamra2023multimodal} incorporated multimodal information and removed some samples with missing information, resulting in a final set of 255 videos. However, both studies suffered from relatively small test set sizes, impacting the stability of their test results. Figure \ref{fig:introduction_2} shows some samples of several commonly used datasets. We observe two problems with the existing datasets. Firstly, in the stage of fire development and outbreak, smoke and flame have been very intense, the value of automatic fire detection and alarm is not high, which is not in line with the original intention of early warning and disaster loss reduction. Secondly, the data styles of laboratory ignition scenes and realistic wildfires are quite different. In the SKLFS-WildFire Test Dataset proposed in this paper, we collect real early wildfire data, where the fire is in the initial stage. This ensures that the dataset is more consistent with real-world scenarios of wildfire detection, allowing for a more direct evaluation of the performance of wildfire detection models in practical applications.
\begin{figure*}[t]
	\centering
	\includegraphics[width=0.99\textwidth]{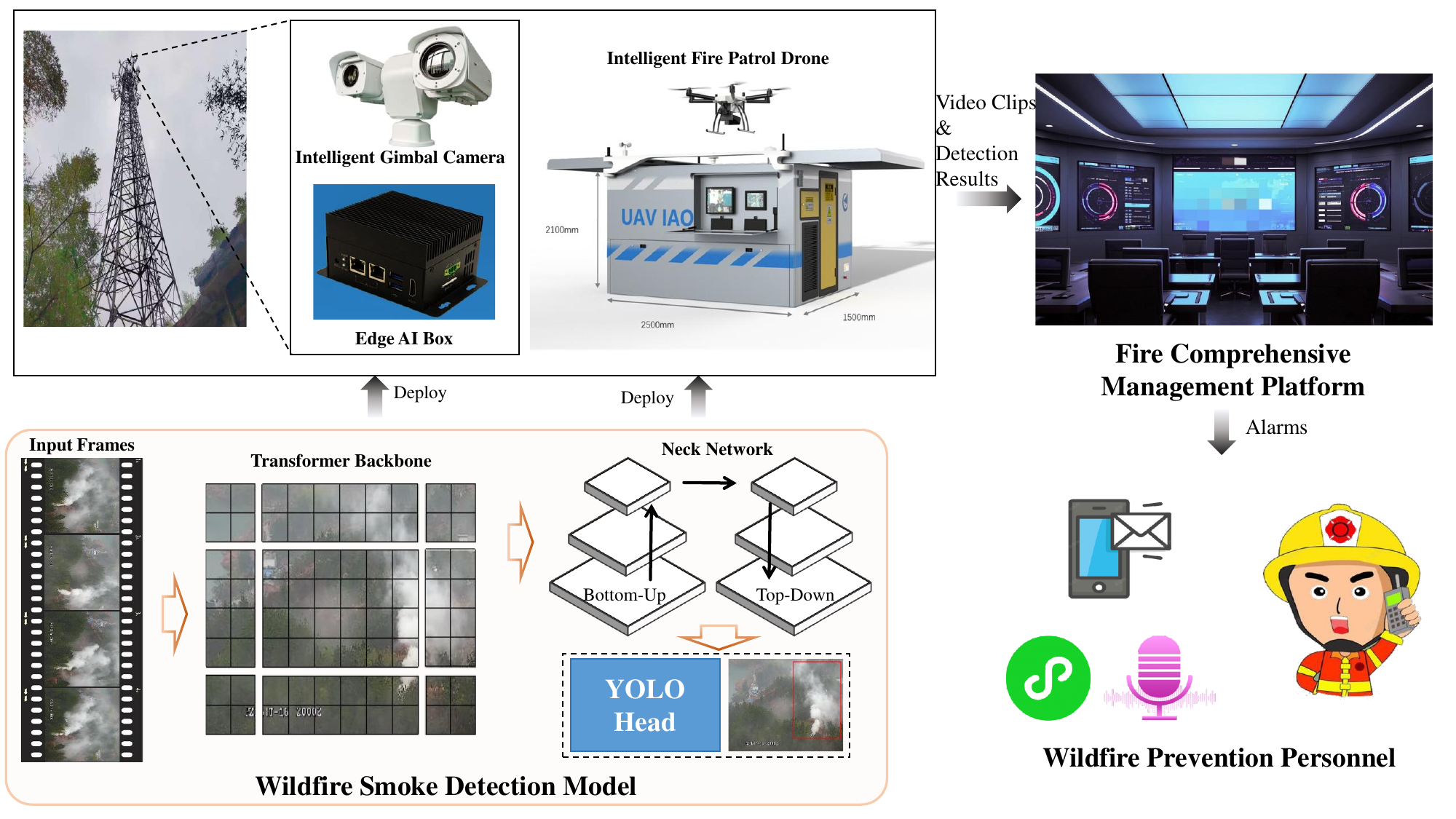}
	\caption{The Wildfire Smoke Detection System. The core of the system is a smoke object detector based on a deep learning model. The model is typically deployed on Intelligent Gimbal Cameras or Edge AI Boxes to perform real-time video analysis, then transmit candidate video segments and detection results back to the Fire Comprehensive Management Platform for confirmation, and finally send confirmed alarms to Wildfire Prevention Personnel.}	
	\label{fig:system}
\end{figure*}
\section{Method}
Wildfire smoke detection is an extension of general object detection. General-purpose object detectors can be generally divided into two categories: single stage and multi-stage. Multi-stage object detectors are generally slightly better than single-stage object detectors due to the feature alignment, which finetunes the proposals from the first stage using the aligned features. However, Figure~\ref{fig:introduction_1} shows that the location and range of smoke are very ambiguous, and the location finetuning in the second stage has little significance, and even degrades the detector performance by introducing excessive location cost. Therefore, based on the classic single-stage detector YOLOX~\cite{ge2021yolox}  , this paper improves the model structure and training strategy according to the particularity of smoke detection. The reason why this paper is not based on more recent studies such as the newest YOLOV8\cite{Jocher_YOLO_by_Ultralytics_2023} is that these methods incorporate a large number of tricks based on general object detection tasks, which are unverified in the field of wildfire detection. Furthermore, the Cross Contrast Patch Embedding(CCPE) and Separable Negative Sampling Mechanism(SNSM) proposed in this paper can be easily applied to the new detector since the conclusion of this paper is general and can be extended.

\subsection{System Framework}
The Wildfire Smoke Detection System is an integral component of forest fire monitoring systems. As illustrated in Figure \ref{fig:system}, the smoke detector constitutes the core of this system. This paper proposes a novel transformer-based smoke detection model from three perspectives: model architecture, training mechanisms, and dataset composition. The proposed model is deployed on intelligent gimbal cameras or edge AI boxes to perform real-time video analysis, transmitting suspicious video segments and detection results back to the Fire Comprehensive Management Platform. Maintenance personnel on duty manually verify whether the detected incidents constitute actual fires. Upon confirmation of high-risk levels, the platform sends alarm signals via text messages, WeChat Mini Programs, voice calls, and other means to notify wildfire prevention personnel for on-site investigation. The system employs a human-machine coupling approach, balancing high recall rates while tolerating a certain number of false alarms. However, an increase in false alarms significantly burdens the workload of personnel.

\subsection{Model Pipeline}

\begin{figure*}[t]
	\centering
	\includegraphics[width=0.99\textwidth]{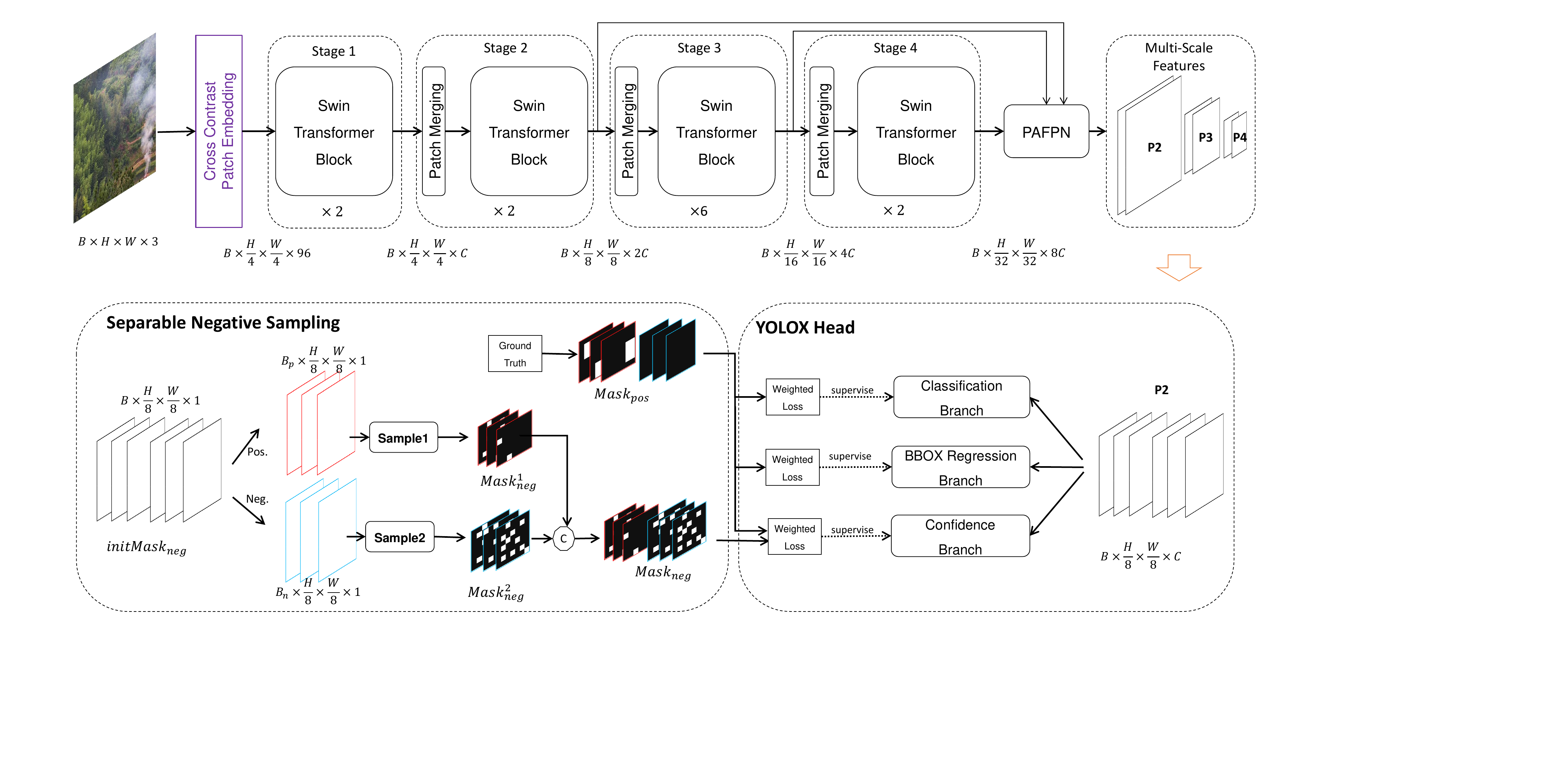}
	\caption{Overall pipeline of the proposed wildfire detector. We use the swin transformer as the backbone network and YOLOX as the detection head. CCPE is used to replace swin's original patch embedding module, and SNSM is used to sample locations involved in training.}	
	\label{fig:pipeline}
\end{figure*}

The overall network structure is shown in Figure \ref{fig:pipeline}. In this paper, the Swin Transformer backbone network is used to extract multi-scale features, the PAFPN \cite{liu2018path} is used for integrating multi-scale features by adding a bottom-up fusion pathway after the top-down fusion pathway of FPN \cite{lin2017feature}. Finally, the YOLOX detection head is used to identify and locate smoke instances.

The input of the backbone network is the RGB image group, denoted as $\{I_i\}_{i=1}^B \in \mathbb{R}^{B\times H\times W\times 3}$, where $B$, $W$, $H$ are the batch size, the width and height of the image respectively. After the backbone network and neck, three scales of features $P2\in \mathbb{R}^{B\times \frac{H}{8}\times \frac{W}{8}\times 2C}$, $P3\in \mathbb{R}^{B\times \frac{H}{16}\times \frac{W}{16}\times 4C}$ and $P4\in \mathbb{R}^{B\times \frac{H}{32}\times \frac{W}{32}\times 8C}$ are obtained, where $C$ is a hyperparameter, refers to output channels of Swin Transformer's first stage. Finally, the multi-scale features are fed into the YOLOX head to predict confidence, class, and bounding box. Transformer has strong modeling ability for context correlation and long-distance dependence, but its ability to capture the detailed texture information of images is weak. To address this problem, this paper redesigns the Patch Embedding of Swin Transformer for somke detection. In order to solve the problem of ambiguity of label assignment caused by the difficulty of determining the smoke boundary, a separate negative sampling mechanism is added to the loss function, and the obtained positive and negative sample masks are used to weight the costs of classification, regression and confidence.

\subsection{Cross Contrast Patch Embedding}

Smoke has a very unique nature, that is, an important clue to the presence of smoke lies in the contrast of color and transparency in space. Capturing spatial contrast can highlight smoke foreground in the background and distinguish fire smoke from homogeneous blurriness, such as poor air visibility, motion blur, zoom blur, and so on. However, the Transformer architecture is weak in capturing low-level visual cues. To solve this problem, we propose a novel Cross Contrast Patch Embedding module, which is composed of a horizontal contrast component and a vertical contrast component in series.

\begin{figure*}
	\begin{minipage}[b]{0.59\linewidth}
		\centering
		%\subfloat[][]{\includegraphics[width=0.9\linewidth,height=3.5cm]{figure_3_1.pdf}}
		\subfloat[][Horizontal Contrast.]{\includegraphics[width=0.98\linewidth]{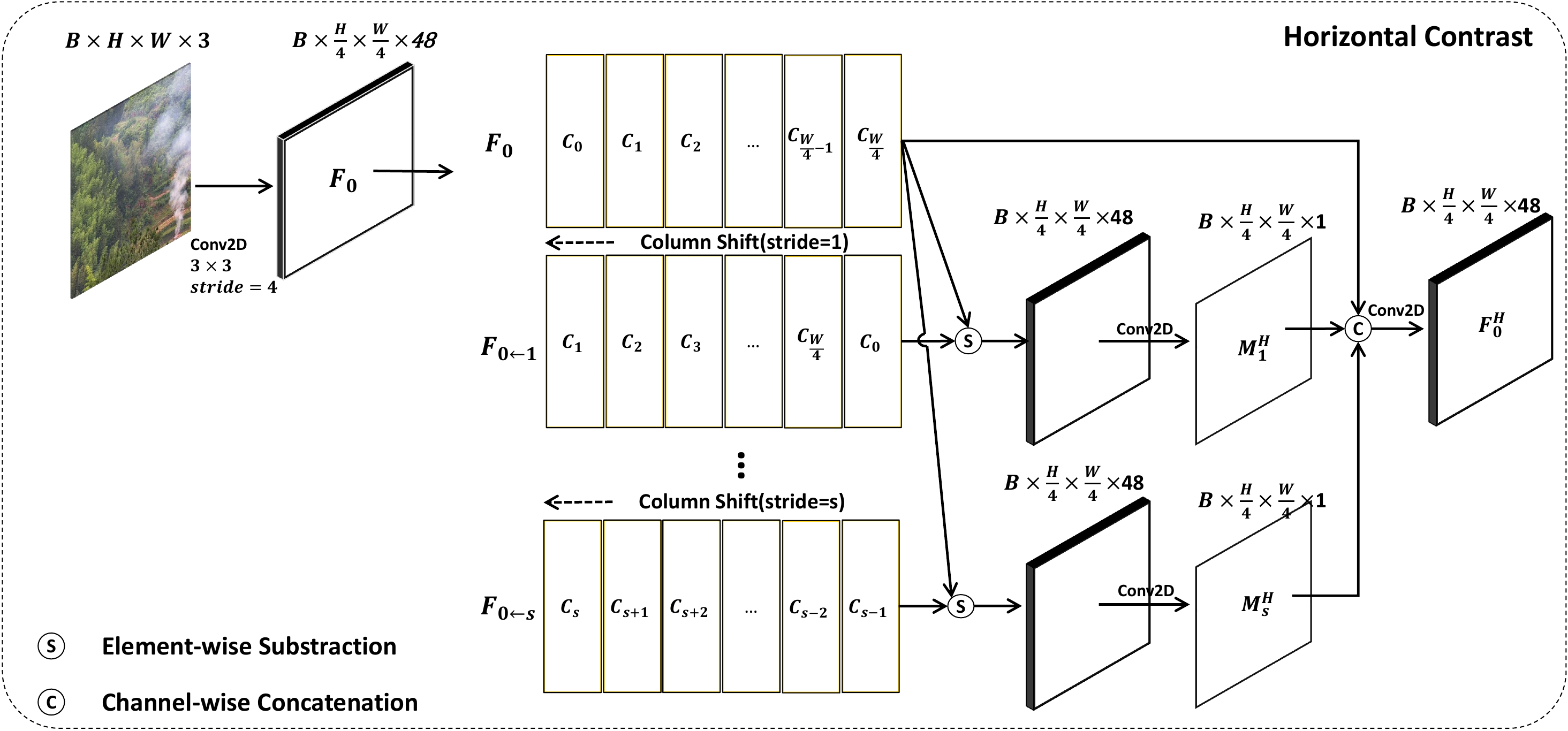}\label{fig:4a}}
	\end{minipage}
	\begin{minipage}[b]{0.40\linewidth}
		\centering
		\subfloat[ ][Vertical Contrast.]{\includegraphics[width=0.98\linewidth]{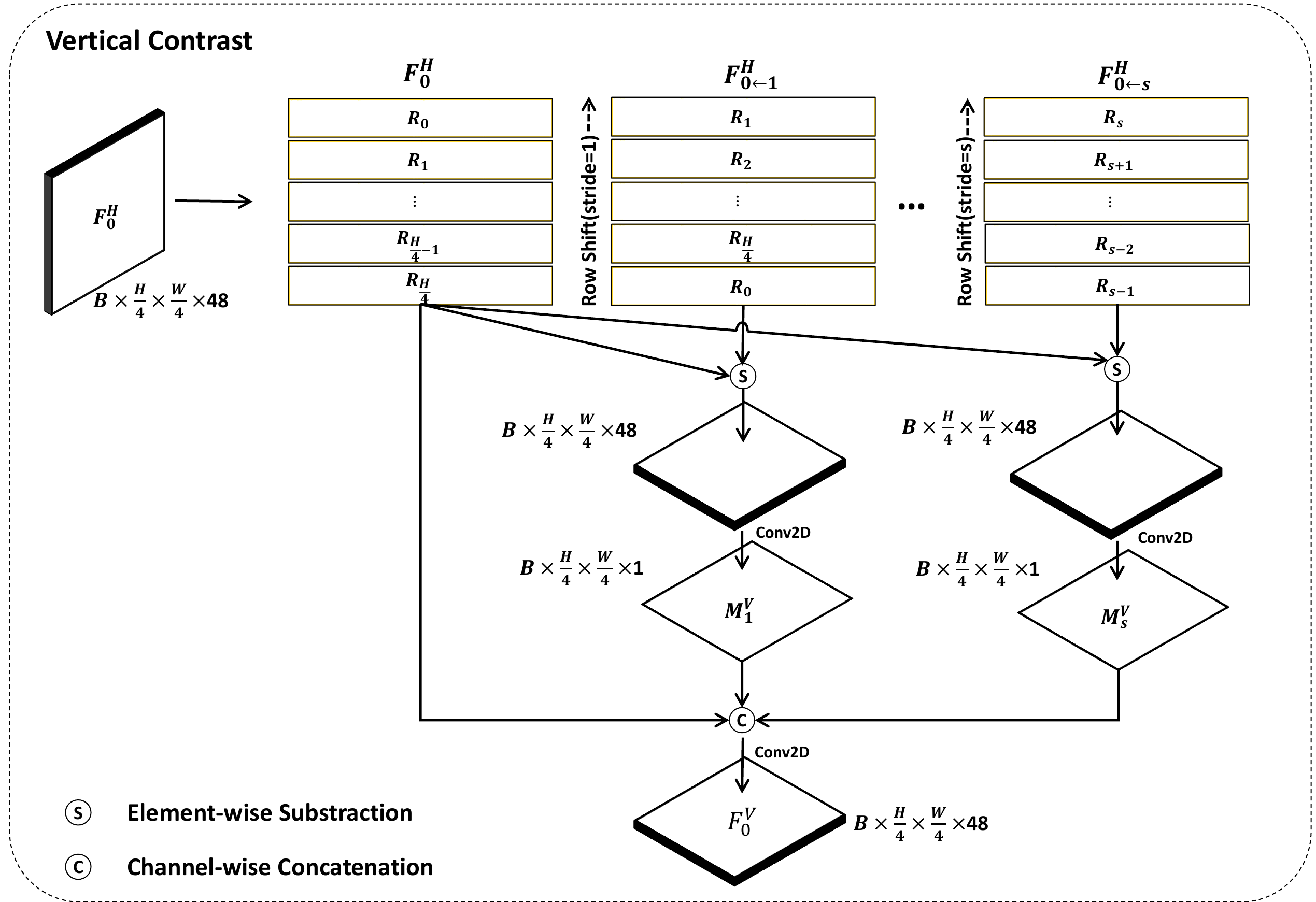}\label{fig:4b}}
	\end{minipage}
	\caption{Demonstration of the CCPE’s two components: a) Horizontal contrast is used to capture the horizontal multi-scale contrast information by using column-wise dislocation subtraction. b) Vertical contrast is connected in series after horizontal contrast, and the use of row-wise dislocation subtraction captures the vertical multi-scale contrast information.
%Demonstration of the CCPE's two components. Horizontal Contrast is used to capture the horizontal multi-scale contrast information by using the column-wise dislocation subtraction. Vertical Contrast is connected in series after Horizontal Contrast, and the vertical multi-scale contrast information is captured by the use of row-wise dislocation subtraction.
}
\label{fig:CCPE}
\end{figure*}

\paragraph{Horizontal Contrast. }
As shown in Figure \ref{fig:4a}, a group of RGB images $\mathscr{I} = \{I_i\}_{i=1}^B \in \mathbb{R}^{B \times H  \times  W  \times  3}$ is fed into a 2D convolution layer with stride 4, and output the feature $F \in \mathbb{R}^{B\times \frac{H}{4}\times \frac{W}{4}\times 48}$. Decompose $F$ into column vectors $\{C_i\}_ {i=0}^{\frac{W}{4}-1}$, and then shift the column to the left. For example, when the shift stride is $s$, a new feature maps is obtained:
\begin{align}
& F_s[j]=F[mod (j+s, \frac{W}{4}],
\end{align}
in which, $[\cdot]$ is the column index operator, and $mod(\cdot)$ is the remainder function. The proposed Horizontal Contrast component contains Column shifts with multiple stride, denoted as the set of strides $\mathscr{S}^H$, and the total number of elements is $S$. After subtracting the misalignment feature $F_s$ generated by each stride from the original feature $F$, a 2D convolution with stride of $3 \times 3$ is performed to obtain the lateral contrast mask, denoted as $ M_s^H$:
\begin{align}
M_s^H=Conv2D(F-F_s).
\end{align}
Finally, the original feature $F$ and the contrast masks of all scales are concatenated on the channel, and a $3\times 3$ 2D convolution is applied to adjust the feature maps $F^H$ to $48$ channels.
\paragraph{Vertical Contrast. }Vertical Contrast does exactly the same thing as Horizontal Contrast, except changing the input to $F^H$, the output of Vertical Contrast, and changing column shift to row shift. Specifically, the input feature map $F^H$ is decomposed into row vectors $\{R_i\}_{i=0}^{\frac{H}{4}-1}$. The set $\mathscr{S}^V$ containing various row shift strides is designed, and then shift the $F^H$ with $s\in \mathscr{S}^V$ as stride to obtain a new feature map:
\begin{align}
\label{eq:loss}
& F_s^H[j]=F^H[mod (j+s, \frac{H}{4}],
\end{align}
in which, $[\cdot]$ is the row index operator. Then, the shifted feature maps are subtracted from $F^H$ respectively, and the potential vertical contrast mask, $M_s^V$, is output after 2D convolution. All masks are concatenated with the $F^H$, and the 48 channel feature map $F^V$ is obtained after a 2D convolution adjustment.

Finally, $F$ and $F^V$ are concatenated in the channel dimension to obtain the final Patch Embedding result. Compared with the vanilla Patch Embedding, the proposed CCPE can quickly capture the multi-scale contrast of smoke images. Moreover, the computational cost of spatial shift operation is very small, only a small amount of convolution operators increases the computational cost. Therefore, CCPE makes up for the natural defects of Transformer in smoke detection, and obtains significant improvement in effect at a controllable computational cost.

\subsection{Separable Negative Sampling}
In this paper, the images with smoke objects are called positive image samples, and the images without smoke are called negative image samples. And, we refer to image regions containing smoke according to detector rules (e.g., IoU threshold) as positive instance samples, while regions without smoke as negative instance samples. 

Smoke detection applications in open scenes often suffer from missed detection and false detection. It has been shown in Figure~\ref{fig:introduction_1} that the smoke boundary is difficult to define. One of the reasons for smoke missed detection is that in the positive image samples, a large number of image regions that may have smoke but are assigned negative labels during the training process, while only the manually labeled regions are assigned positive labels. The areas with smoke but get negative labels often contribute more loss, which increases the difficulty of training, and then leads to a high false negative rate of the detector. Meanwhile, the cause of smoke false detection lies in the complex background in open scenes. There are a large number of image regions with high appearance similarity to smoke. In engineering, false detections can be suppressed by adding error-prone images. The error-prone region only accounts for a small part of the image, so the loss share of the error-prone region can be strengthened by Online Hard Example Mining (OHEM)~\cite{shrivastava2016training}. OHEM involves identifying the top-K negative samples with the highest scores during training, which typically represent hard-negative examples. If the traditional OHEM is used to highlight the difficult regions on all the images, the label ambiguity problem in the positive image samples will be more prominent. Therefore, we propose the Separable Negative Sampling Mechanism (SNSM) to alleviate this problem.

As shown in Figure~\ref{fig:pipeline}, the feature maps $P2\in \mathbb{R}^{B\times \frac{H}{8}\times \frac{W}{8}\times 2C}$ is taken as an example. YOLOX head has three branches for confidence prediction, classification and regression respectively. The positive mask $mask_{pos}$ is determined using manually annotated Ground Truth as well as positive sample assignment rules. YOLOX sets the center region of $3\times 3$ as positive, and this paper follows this design. The classification loss as well as the loss for the regression are weighted using $Mask_{pos}$. In the vanilla YOLOX head, all spatial locations contribute confidence loss. However, due to the addition of a large number of negative error-prone images, we use all positive locations, namely $Mask_{pos}$, and part of the negative locations after sampling to weight the confidence loss by location.

Considering the importance of sampling negative locations for loss computation, we propose the separable negative sampling mechanism. Denote the initial mask containing all negative locations as $initMask_{neg} =1-mask_{pos}$. Divide $initMask_{neg}$ into two groups according to the presence or absence of smoke, That is, the positive image group $initMask_{neg}^1 \in \mathbb{R}^{B_p\times \frac{H}{8}\times \frac{W}{8}\times 2C}$ and the negative image group $initMask_{neg}^2\in \mathbb{R}^{B_n\times \frac{H}{8}\times \frac{W}{8}\times 2C}$, where$B_p$ and $B_n$ are the number of positive and negative images, respectively. In $initMask_{neg}^1$, negative locations are randomly sampled according to the number of positive locations, and the result is denoted as $mask_{neg}^1$. We randomly select a number of negative samples that is $\alpha ^1$ times the number of positive samples.
In $initMask_{neg}^2$, OHEM is used to collect negative samples. Initially, all negative samples are ranked in descending order based on their scores, and then the top-$K$ negative samples are selected for training, which are denoted as $mask_{neg}^2$. Here, $K$ is set to $\alpha^2$ times the number of positive samples. We set $\alpha ^1 \gg \alpha^2$, so that the negative samples learned by the model are more from negative images, and more attention is paid to the image regions prone to misdetection due to OHEM.

\section{Experiments}
\subsection{Datasets}
\textbf{FIgLib} \cite{dewangan2022figlib} introduced by Dewangan \emph{et al.} in 2022, stands as the largest forest fire dataset to date. It encompasses 315 wildfire videos, of which 270 are usable. The dataset is divided into three subsets: training, validation, and testing, comprising 144, 64, and 62 videos, respectively. Notably, FIgLib's distinguishing feature lies in its reliance on temporal information for smoke identification. Even human intelligence struggles to discern smoke solely from single-frame images within FIgLib.

\textbf{SKLFS-WildFire} is an open scene early fire dataset, which contains two parts: training set and test set. Train Split contains 299,827 images derived from 30,470 video clips, of which 72,145 images contain smoke objects. The Test Split is all derived from real forest fire videos and contains 50,735 images (form 3,309 videos), of which 3,588 images are positive images (from 340 videos). There are two image sizes: $2688\times1520$ and $1920\times1080$, and all the videos were captured from visible light cameras of unknown make and model. SKLFS-WildFire has 4 characteristics:
\begin{enumerate}
\item{High diversity.} Most of the data of SKLFS-WildFire comes from video, but we're not taking too many images on one video. About 10 frames are extracted from each video on average to ensure the diversity of data.
\item{Strict Train-Test split design.} We split the test dataset according to the administrative region where the cameras are deployed, and all the test videos are in completely different geographical locations from the train dataset.
\item{Difficult negative samples.} We added a large amount of non-smoke data on both the training and test sets. These non-smoke data come from the false detection of classical detectors such as Faster R-CNN \cite{ren2016faster}, YOLOV3 \cite{redmon2018yolov3} and YOLOV5 \cite{jocher2022ultralytics}.
\item{Realistic early fire scenes.} All SKLFS-WildFire data comes from the backflow of application data, mainly for the initial stage of fire, not the ignition data in the laboratory scenario, nor the middle and late stage of wildfire.
\end{enumerate}
To protect user privacy, the training set of SKLFS-WildFire is not available for the time being. We open up the SKLFS-WildFire Test to facilitate academic researches.
\subsection{Metrics for SKLFS-WildFire}
The biggest difference between smoke detection and general object detection is the ambiguity of the boundary. Therefore, directly using the commonly used metrics in object detection cannot comprehensively reflect the quality of the model. Therefore, we evaluate the comprehensive performance of the model from the three levels: bounding box, image, and video. At the bounding box level, we use the PR curve and the Average Precision with the IoU threshold of 0.1 ($AP@0.1$). At the image/video level, we treat the smoke detection as a classification task and use PR curve and ROC curve as well as AUC to evaluate the image/video classification performance. It should be pointed out that we take the maximum score of all bounding boxes in a image/video as the score of whether a image/video exists smoke or not. We compute AUC using the Mann Whitney U Test:
\begin{align}
AUC = \frac{\sum_{i,j}I(Score_i, Score_j)}{|\mathscr{D}_{pos}|*|\mathscr{D}_{neg}|},i\in \mathscr{D}_{pos}, j\in \mathscr{D}_{neg},
\end{align}
in which, $\mathscr{D}_{pos}$ and $\mathscr{D}_{neg}$ represent the set of positive and negative respectively. $Score_i$ is the sample $i$'s score. $|\cdot|$ is the cardinal number operator, the $I(\cdot)$ is the scoring function defined as follow:
$$
I(A, B) = \left\{
\begin{array}{rcl}
1       &if \quad      & { A > B}\\
0.5     &if \quad       & {A = B}\\
0     &if \quad       & {A < B} 
\end{array} \right. 
.
$$

\begin{table*}
\begin{center}
\caption{Comparative Results on the FIgLib Dataset. The proposed model demonstrates strong performance in single-frame input models and achieves the best results in multi-frame input models on the FIgLib dataset.}
\renewcommand\arraystretch{2.0}
\resizebox{0.99\textwidth}{!}{
\begin{tabular}{l|l|c|c|c|c|c|c}
\hline

Models   & Backbone                       &   Params(M)      &  \quad  Acc(\%) \quad  & \quad  F1(\%) \quad  &\quad  Precision(\%) \quad &\quad Recall(\%)\quad &\quad TTD\quad  \\
\hline
\multicolumn{8}{l}{\textbf{Single-frame models}}\\
\hline \hline
OURS&   ContrastSwin      & 35.9 &  81.48        &    81.12      &    83.74     &     78.66         &  2.73             \\
\makecell[l]{SmokeyNet \cite{dewangan2022figlib}}&ResNet34+ViT        & 40.3 &  \textbf{82.53}    & \textbf{ 81.30}     & \textbf{ 88.58}      &    75.19         &  2.95                  \\
\makecell[l]{SmokeyNet \cite{dewangan2022figlib}}& ResNet34    & 22.3 &   79.40    &  78.90     &  81.62      &     76.58         &  2.81                  \\
\makecell[l]{SmokeyNet \cite{dewangan2022figlib}} &ResNet50      & 26.1 &  68.51     &  74.30     &  63.35      &    \textbf{89.89 }         &  \textbf{1.01}                  \\
\makecell[l]{FasterRCNN \cite{dewangan2022figlib}}  &ResNet50     & 41.3 &  71.56     &  66.92     &  81.34      &    56.88          &  5.01                  \\
\makecell[l]{MaskRCNN \cite{dewangan2022figlib}}    &ResNet50     & 43.9 &  73.24     &  69.94     &  81.08      &    61.51         &  4.18                  \\
\hline
\multicolumn{7}{l}{\textbf{Multi-frame models} }\\
\hline \hline
OURS&  2 frames ContrastSwin         &35.9   &  \textbf{84.67}&\textbf{84.05} &  88.74      &\textbf{79.83} &\textbf{2.26}    \\
\makecell[l]{SmokeyNet \cite{dewangan2022figlib}}&\makecell[l]{3 frames ResNet34 +\\ LSTM + ViT}& 56.9  & 83.62     & 82.83     &  \textbf{90.85}&  76.11          &  2.94               \\
\makecell[l]{SmokeyNet \cite{dewangan2022figlib}}&\makecell[l]{2 frames ResNet34 + \\ LSTM + ViT }       & 56.9  &  83.49     &  82.59     & 89.84       & 76.45            &  3.12                  \\
\makecell[l]{SmokeyNet \cite{dewangan2022figlib}}&\makecell[l]{2 frames MobileNet+\\ LSTM + ViT}      & 36.6  &   81.79     & 80.71    &   88.34     &   74.31           &   3.92                 \\
\makecell[l]{SmokeyNet \cite{dewangan2022figlib}}&\makecell[l]{2 frames TinyDeiT +\\ LSTM + ViT}        & 22.9  &  79.74    & 79.01     &   84.25    &   74.44         &     3.61             \\
\hline
\end{tabular}
}
\label{tab:fIglib}
\end{center}
\end{table*}

\subsection{Implement Details}
All models are implemented using the MMDetection~\cite{mmdetection} framework based on the PyTorch~\cite{imambi2021pytorch} backend. We used the NVIDIA A100 GPUs to train all models while using a single NVIDIA GeForce RTX 3080Ti to compare inference efficiency. The input of the wildfire smoke detection models is a batch of images, and we set the batch size $B$ to $64$. In the training phase, we follow the data augmentation strategy of YOLOX and use mosaic, random affine, random HSV, and random flip to enhance the diversity of data and improve the generalization ability of the models. Finally, through resize and padding operations, the size $W$ and $H$ of the input are both 640. The set of horizontal strides $S^H$ and the set of vertical strides $S^V$ in CCPE are both set as $\{1,2,4,8,16,32,64,128\}$ to capture the contrast information of different scales with eight kinds of strides. In SNSM, the negative sampling ratio $\alpha^1$ of positive image samples is set to $10$, and the negative sampling ratio $\alpha^2$ of negative image samples is set to $190$. The proposed model is optimized with the SGD optimizer with an initial learning rate of $0.01$, momentum set to $0.9$ and weight decay set to $0.0005$, and a total of $40$ epochs are trained. The mixed precision training method of float32 and float16 was used in the training, and the loss scale was set to $512$.

\subsection{Comparison on FIgLib}
We did not find bounding box-level labels for FIgLib, so we reannotated the boxes in the training subset. Since smoke from FIgLib is visually more challenging to discern, we increased the input size of the images from 640 to 1024. Due to FIgLib's emphasis on temporal features for smoke detection, we not only retrained the proposed image model on its training set but also concatenated the current frame $I_t$ with the past frame $I_{t-2}$ in the image channels to obtain a temporal model, where $t$ and $t-2$ represent the current time and the time two minutes ago (the frame interval in FIgLib is one minute). Although designing a more refined temporal model is possible, it is beyond the scope of this paper. We adopt the metrics in SmokeyNet \cite{dewangan2022figlib} and compare the results in P (precision), R (recall), F1 score, Acc (accuracy), and TTD (Time To Detection, minutes) against those reported in \cite{dewangan2022figlib}. Accuracy and F1 score are the primary composite metrics of importance, with other metrics serving as supplementary references. Table \ref{tab:fIglib} presents the comparison of the effects. In the context of single-frame input models, the proposed model has achieved comparable results, although not the best. This is attributed to the smoke in the FIgLib dataset, which is prone to being confused with other objects in single-frame images, thereby hindering more expressive models from achieving superior performance. A testament to this observation is the comparison between the results of ResNet34 in row 5 and ResNet50 in row 6. Conversely, in the realm of multi-frame input models, even with our model's simple utilization of Concatenation to model temporal sequences, we have still attained state-of-the-art (SOTA) results. This underscores the efficacy of the proposed approach.

\subsection{Ablation Studies}
The Cross Contrast Patch Embedding module is proposed to make up for the insufficient ability of the Transformer backbone network to capture the details of smoke texture changes in space. A separable negative sampling strategy is proposed to alleviate the ambiguity of smoke negative sample assignment and improve the recall rate, that is, the model performance under the premise of high recall. In this section, the effectiveness of these two methods is verified by ablation experiments on the SKLFS-WildFire Test dataset.

\begin{table*}
\begin{center}
\caption{Effectiveness of Cross Contrast Patch Embedding on Swin. CNN-based YOLOX significantly outperforms Swin-based YOLOX on BBox-level metric. After using CCPE to improve Swin, model performance significantly exceed the baseline on all metrics.}
\renewcommand\arraystretch{1.4}
\resizebox{0.99\textwidth}{!}{
\begin{tabular}{l|l|c|c|c||c|c}
\hline
Models       &   Backbone \quad\quad\quad\quad\quad\quad     &  \quad  BBox $AP@0.1$ \quad   &  \quad  Image AUC \quad  & \quad Video AUC \quad & \quad Params (M)\quad &\quad GFLOPs\quad \\
\hline\hline
YOLOX       &   CSPDarknet Large      & 0.506       & 0.712      & 0.900   & 54.209 & 155.657 \\
YOLOX-Swin       &  Swin Tiny      & 0.476       &   0.732    &   \textbf{0.909} & 35.840 & 50.524 \\
YOLOX-ContrastSwin \quad\quad       & CCPE + Swin Tiny      & \textbf{0.537}     &   \textbf{0.765}    &   0.908 & 35.893 & 53.250 \\
%\hline\hline
%YOLOX-ContrastSwin (run2)       & CCPE + Swin Tiny      & 0.556    & 0.783    &   0.911   &    35.893 & 53.250     \\
%YOLOX-ContrastSwin (run3)       & CCPE + Swin Tiny      & 0.579   &  0.781       &  0.922     &    35.893 & 53.250    \\
\hline
\end{tabular}
}
\label{tab:ccpe}
\end{center}
\end{table*}

\begin{figure*}
	\begin{minipage}[b]{1.0\linewidth}
		\flushleft 
		\subfloat[][BBox PR Curve.]{\includegraphics[width=0.32\linewidth]{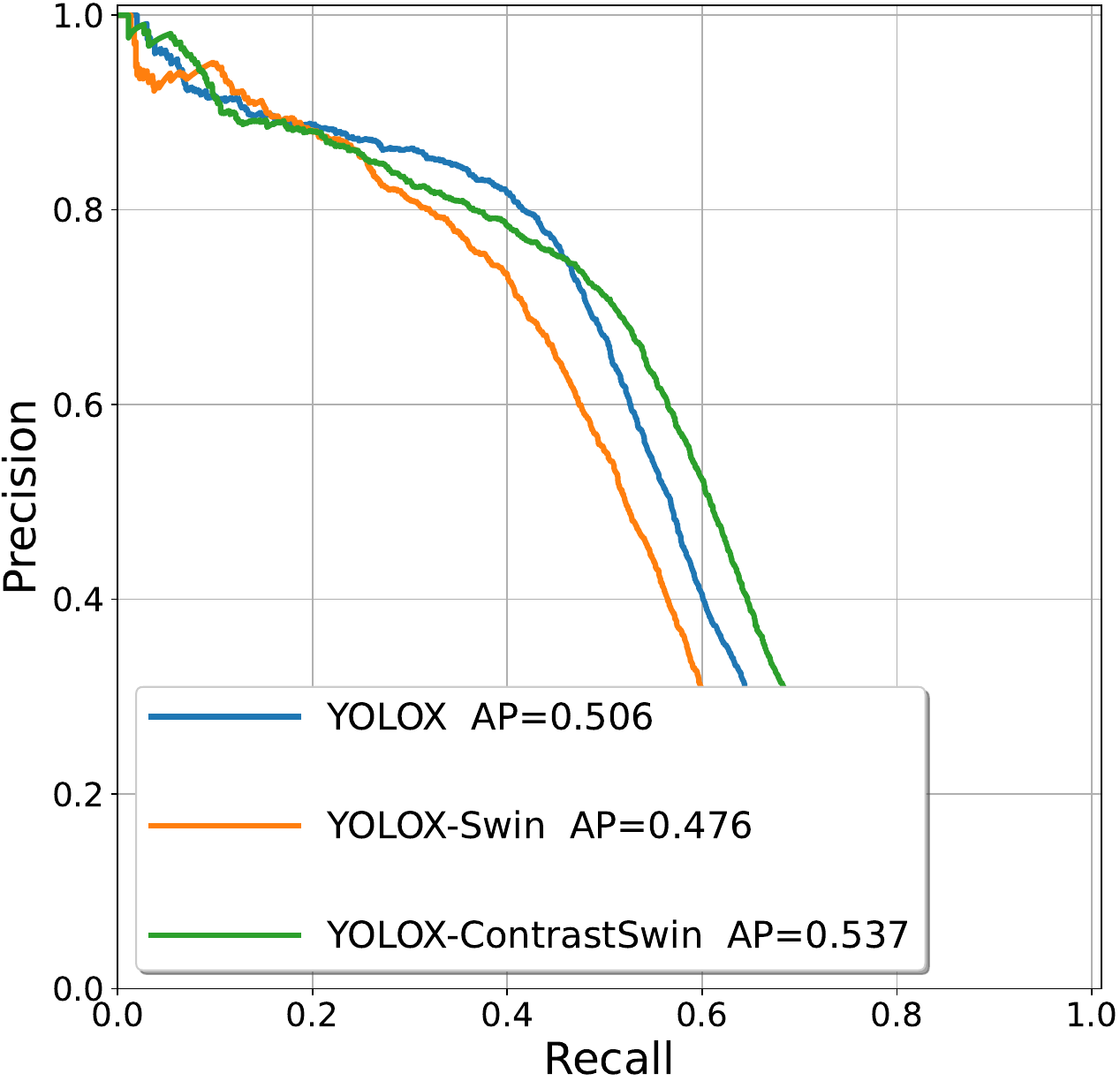}\label{CCPEfigure_1}}		\quad
		\subfloat[][Image PR Curve.]{\includegraphics[width=0.32\linewidth]{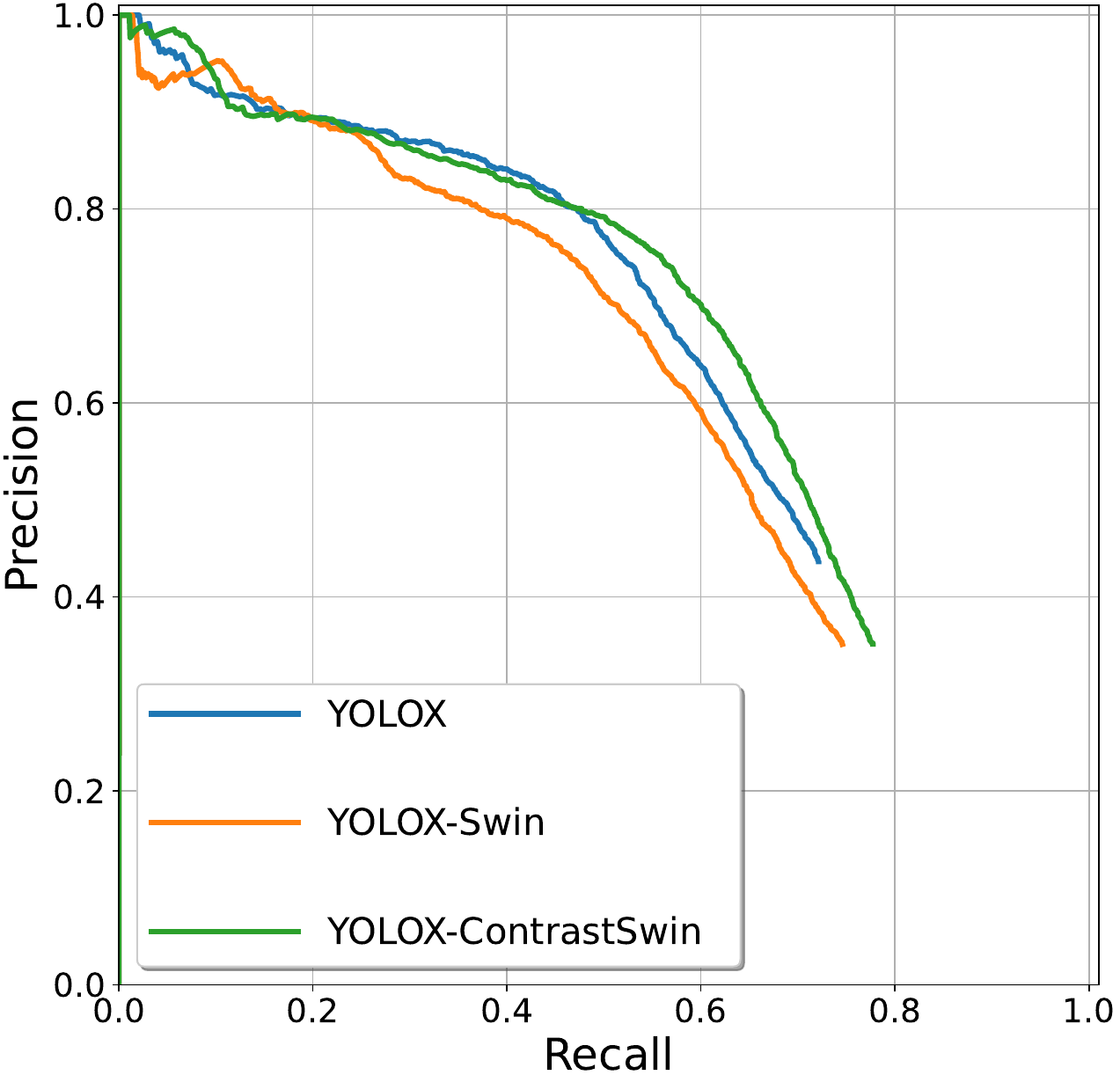}\label{CCPEfigure_2}}\quad
		\subfloat[][Image ROC Curve.]{\includegraphics[width=0.32\linewidth]{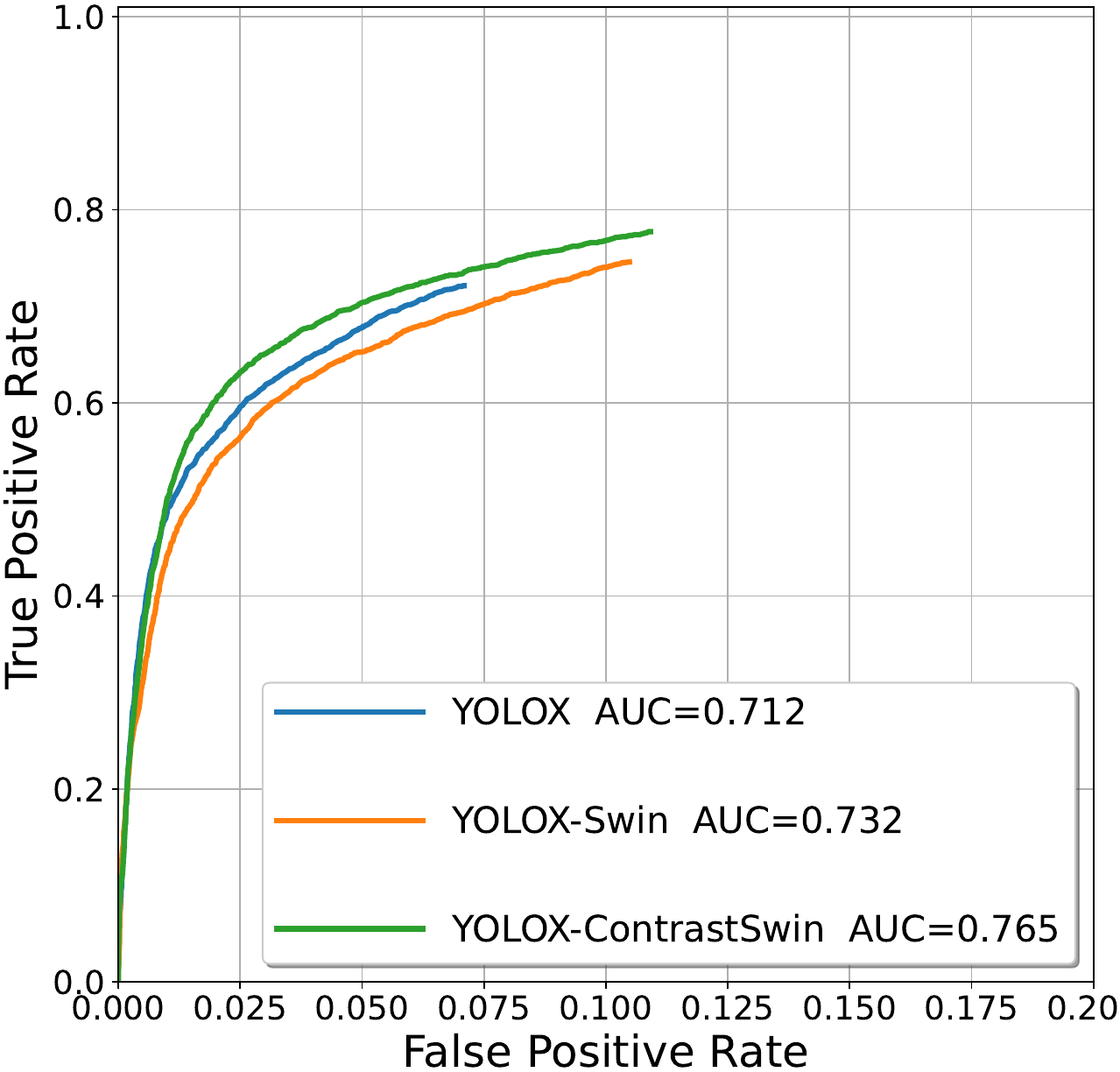}\label{CCPEfigure_3}}\quad
	\end{minipage}
	\begin{minipage}[b]{1.0\linewidth}
		\flushleft 
		\subfloat[][Video PR Curve.]{\includegraphics[width=0.32\linewidth]{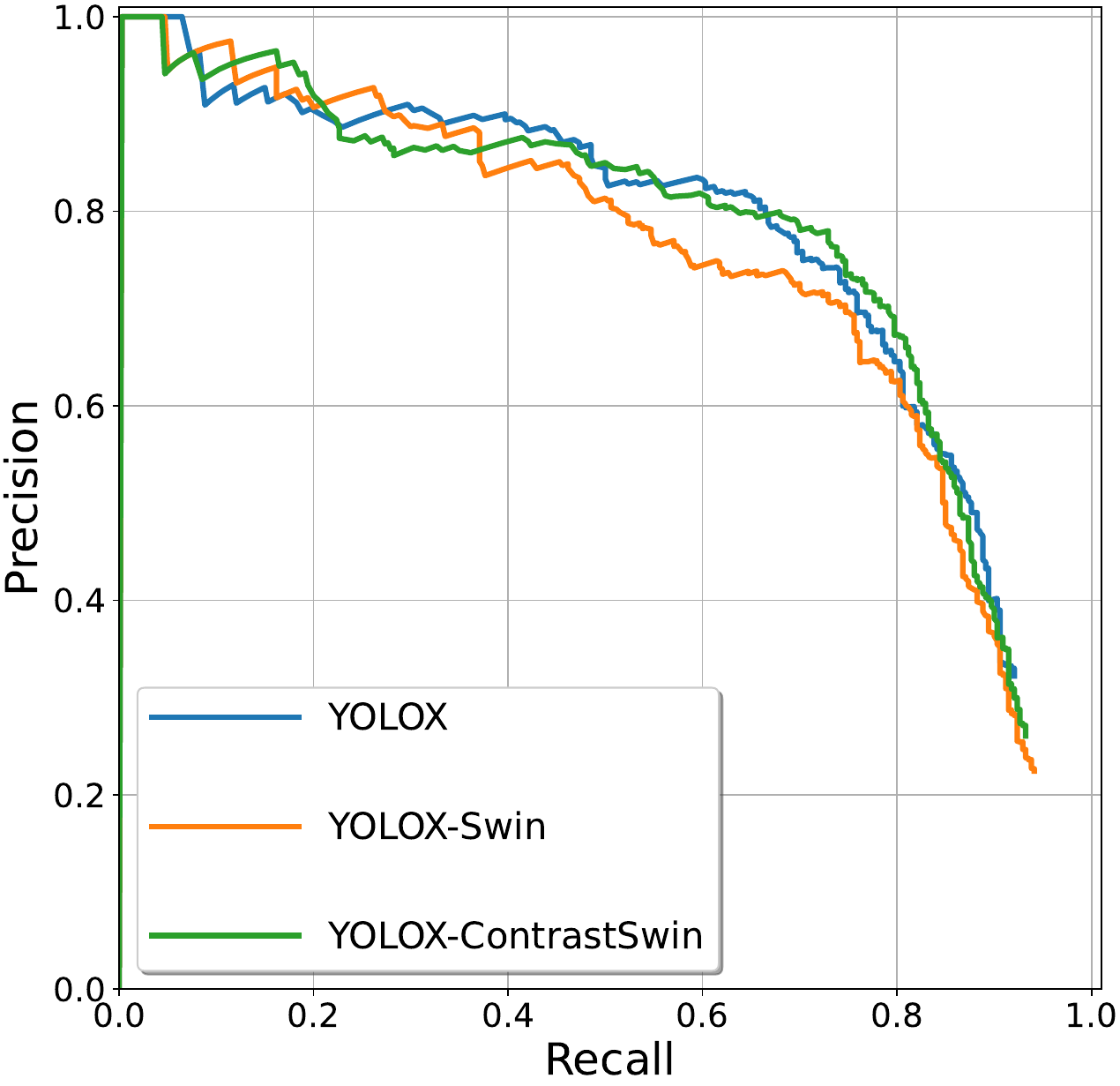}\label{CCPEfigure_4}}\quad
		\subfloat[][Video ROC Curve.]{\includegraphics[width=0.32\linewidth]{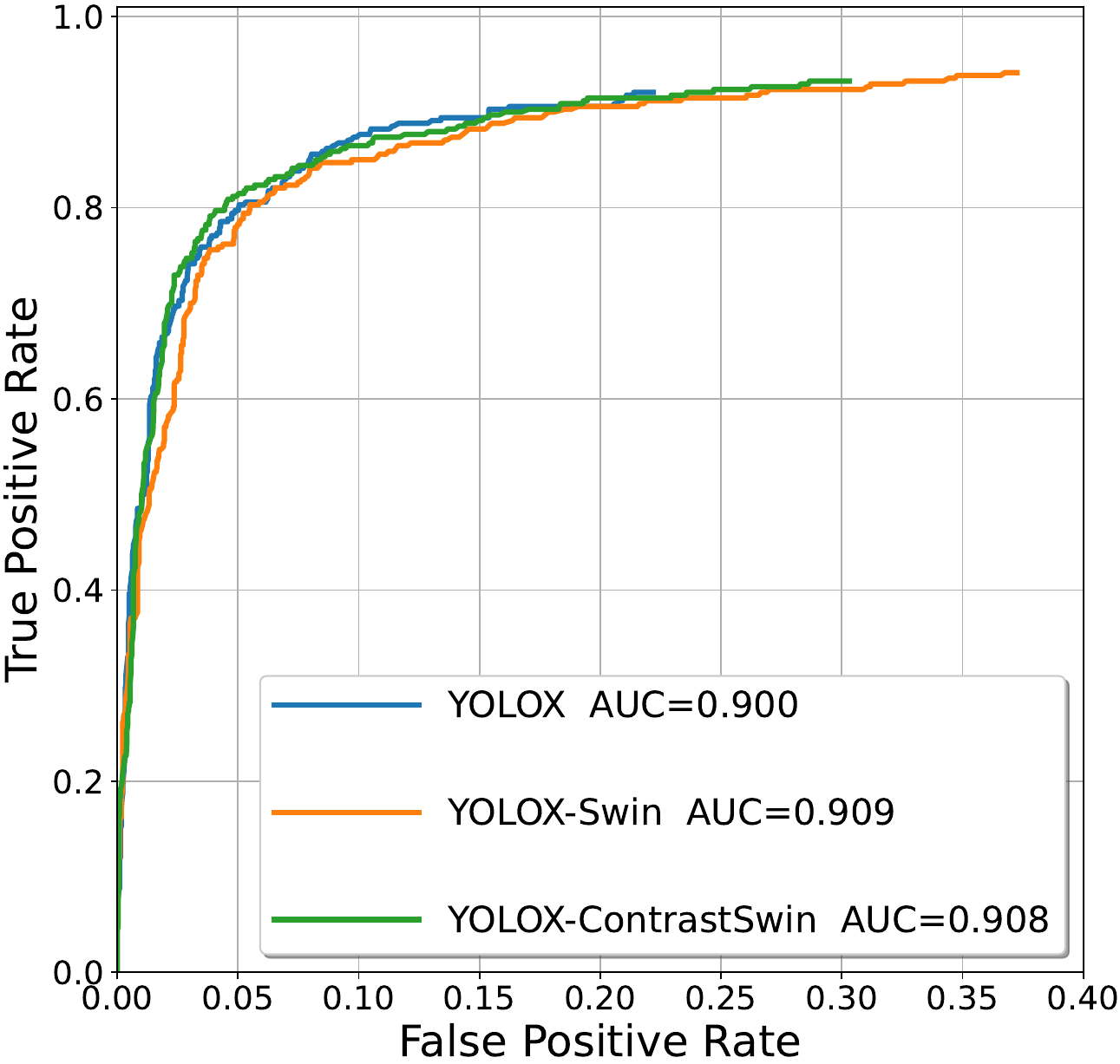}\label{CCPEfigure_5}}\quad		
           \subfloat[][Comparison of numerical metrics.]{\includegraphics[width=0.32\linewidth]{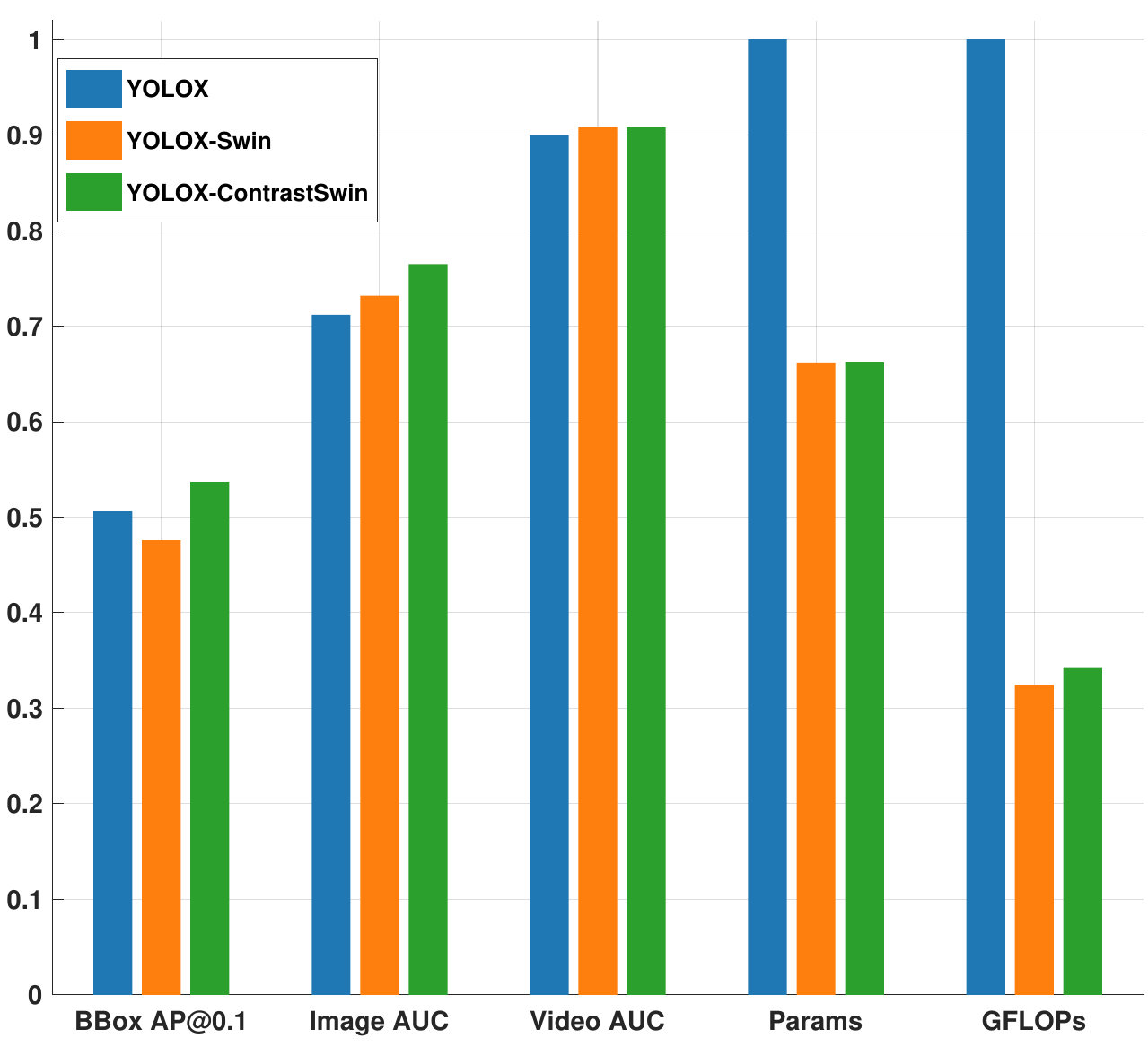}\label{hist_contrast}}
	\end{minipage}
	\caption{Comparison of CCPE and two baseline models. (a),(b),(c),(d) and (e) are bounding box level PR Curve, image lever PR Curve, image level ROC Curve, video lever PR Curve, video level ROC Curve. (f) shows the numerical metrics comparison, where Params and GFLOPs are the normalized results.}
\label{fig:CCPE}
\end{figure*}

\paragraph{Effectiveness of Cross Contrast Patch Embedding.} As shown in the Table \ref{tab:ccpe}, after replacing the CNN-based backbone network CSPDarknet \cite{bochkovskiy2020yolov4} with Swin Transformer Tiny (denoted as Swin-T) in YOLOX, the AP of bounding box level decreased.
 The AUC of image level and video level are significantly improved. This proves our observation that the Transformer architecture has a stronger ability to model contextual relevance, while CNNs are better at capturing the detailed texture information for better location prediction. In the last row of the table, the patch embedding of Swin-T is replaced by the proposed Cross Contrast Patch Embedding. The experimental results show that CCPE consistently outperforms the original YOLOX model and the model replaced by the Swin-T backbone in terms of bounding box, image and video levels. Figure~\ref{fig:CCPE} presents the PR curves at the bounding box level and the PR curves and ROC curves at image and video levels. These curves show the same conclusions as the numerical metrics, proving that the proposed CCPE can effectively make up for the lack of Transformer model ability to model the low-level details of smoke, and significantly improve the comprehensive performance of smoke object detection model with a small number of parameters and calculation.

\begin{table*}
\begin{center}
\caption{Effectiveness of Separable Negative Sampling. Although there is a decrease in box level AP, the proposed SNSM is able to significantly enhance the comprehensive metrics at image level and video level.}
\renewcommand\arraystretch{1.4}
\resizebox{0.99\textwidth}{!}{
\begin{tabular}{l|l|c|c|c||c|c}
\hline
Models       &   Sampling\quad \quad\quad\quad\quad\quad     &    \quad BBox $AP@0.1\quad$    &    \quad Image AUC\quad   &  \quad Video AUC\quad  &  \quad Params (M)\quad & \quad GFLOPs\quad \\
\hline\hline
YOLOX-ContrastSwin \quad\quad&   None(Baseline)      & 0.537     &   0.765    &   0.908 & 35.893 & 53.250 \\
YOLOX-ContrastSwin \quad\quad&  Random      & 0.527        & 0.847      & 0.933   & 35.893 & 53.250 \\
YOLOX-ContrastSwin \quad\quad       & OHEM     & \textbf{0.574}     & 0.783       & 0.924    & 35.893 & 53.250 \\
YOLOX-ContrastSwin \quad\quad       & SNSM(OURS) &  0.503     & \textbf{0.900}       &  \textbf{0.934}   & 35.893 & 53.250 \\
%\hline\hline
%YOLOX-ContrastSwin (run2)       & SNSM(OURS)      & 0.496     &  0.904    &  0.933  &   35.893 & 53.250     \\
%YOLOX-ContrastSwin (run3)       &SNSM(OURS)     &   0.493  &   0.904     &   0.932   &    35.893 & 53.250  \\
\hline
\end{tabular}
}
\label{tab:SNSM}
\end{center}
\end{table*}

\begin{figure*}
	\begin{minipage}[b]{1.0\linewidth}
		\flushleft 
		\subfloat[][BBox PR Curve.]{\includegraphics[width=0.32\linewidth]{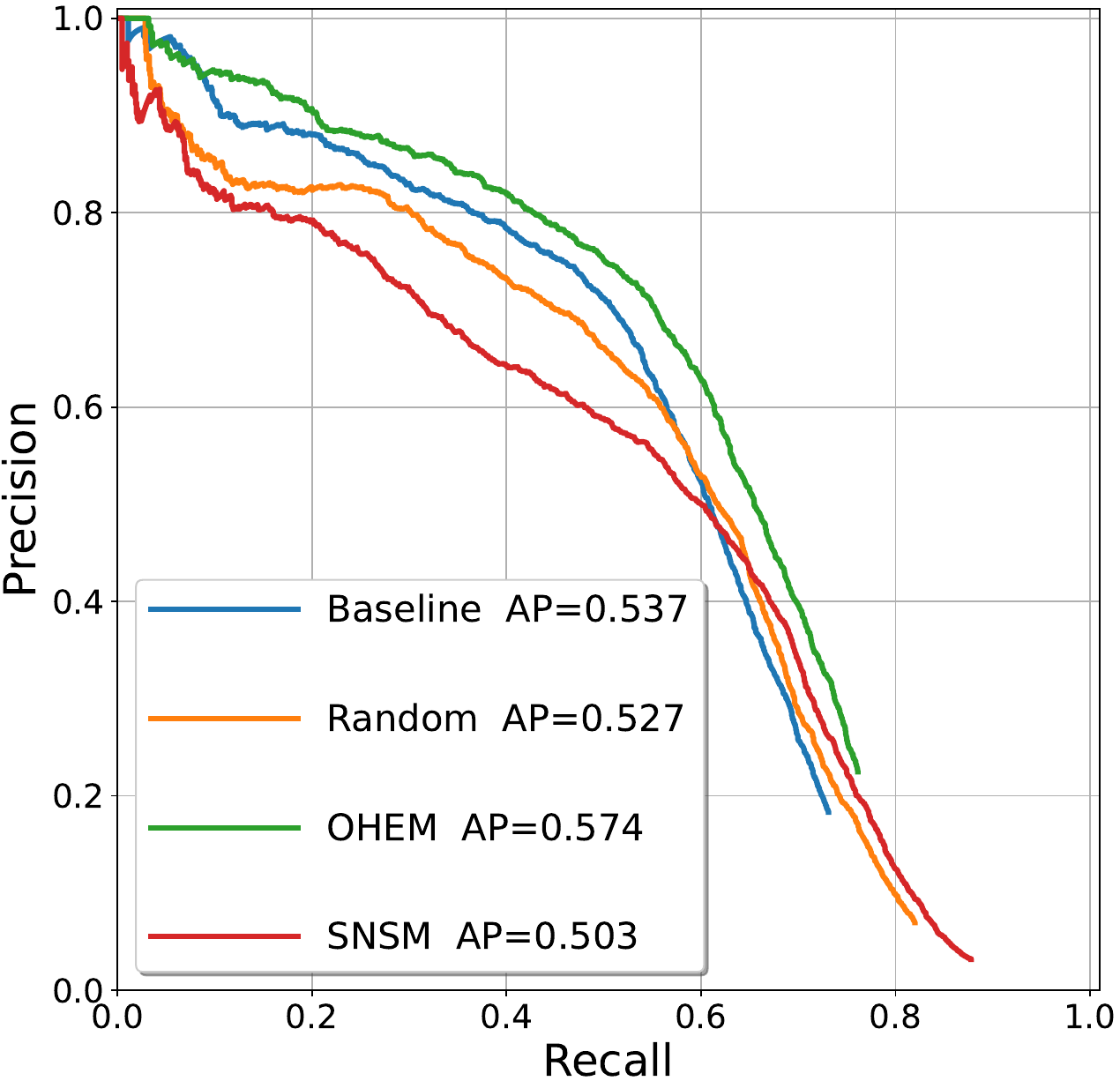}\label{SNSMfigure_1}}		\quad
		\subfloat[][Image PR Curve.]{\includegraphics[width=0.32\linewidth]{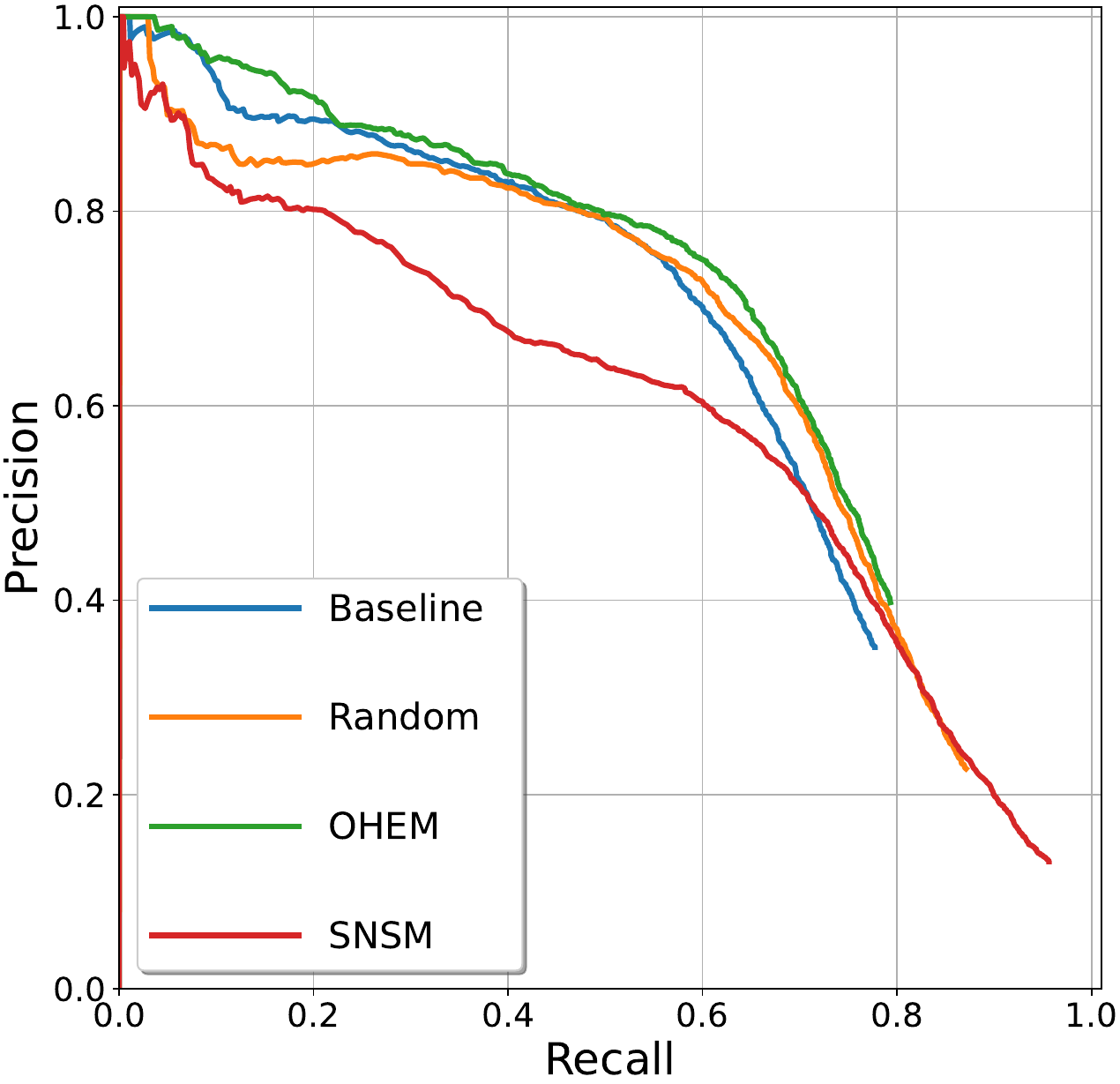}\label{SNSMfigure_2}}\quad
		\subfloat[][Image ROC Curve.]{\includegraphics[width=0.32\linewidth]{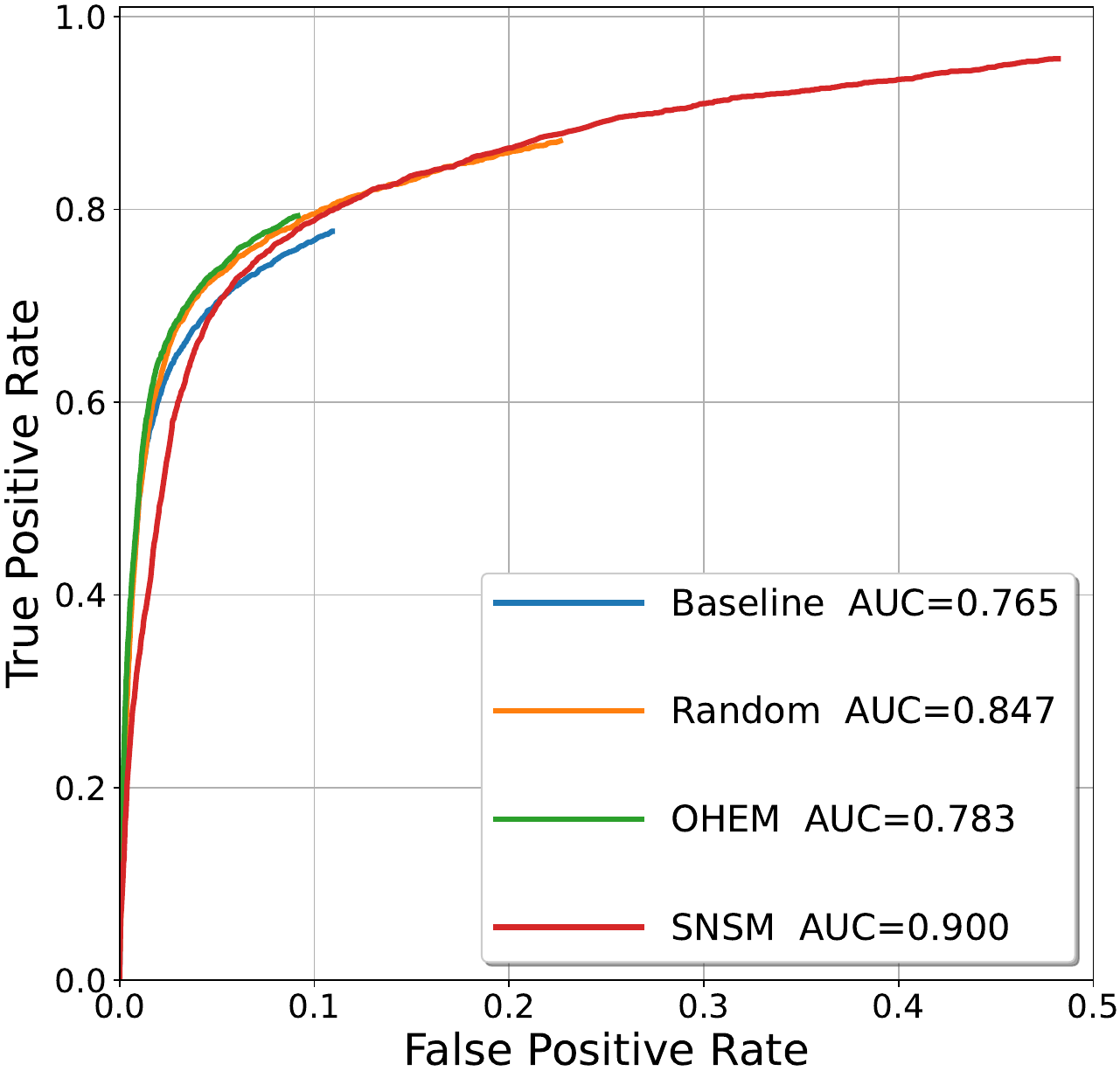}\label{SNSMfigure_3}}\quad
	\end{minipage}
	\begin{minipage}[b]{1.0\linewidth}
		\flushleft 
		\subfloat[][Video PR Curve.]{\includegraphics[width=0.32\linewidth]{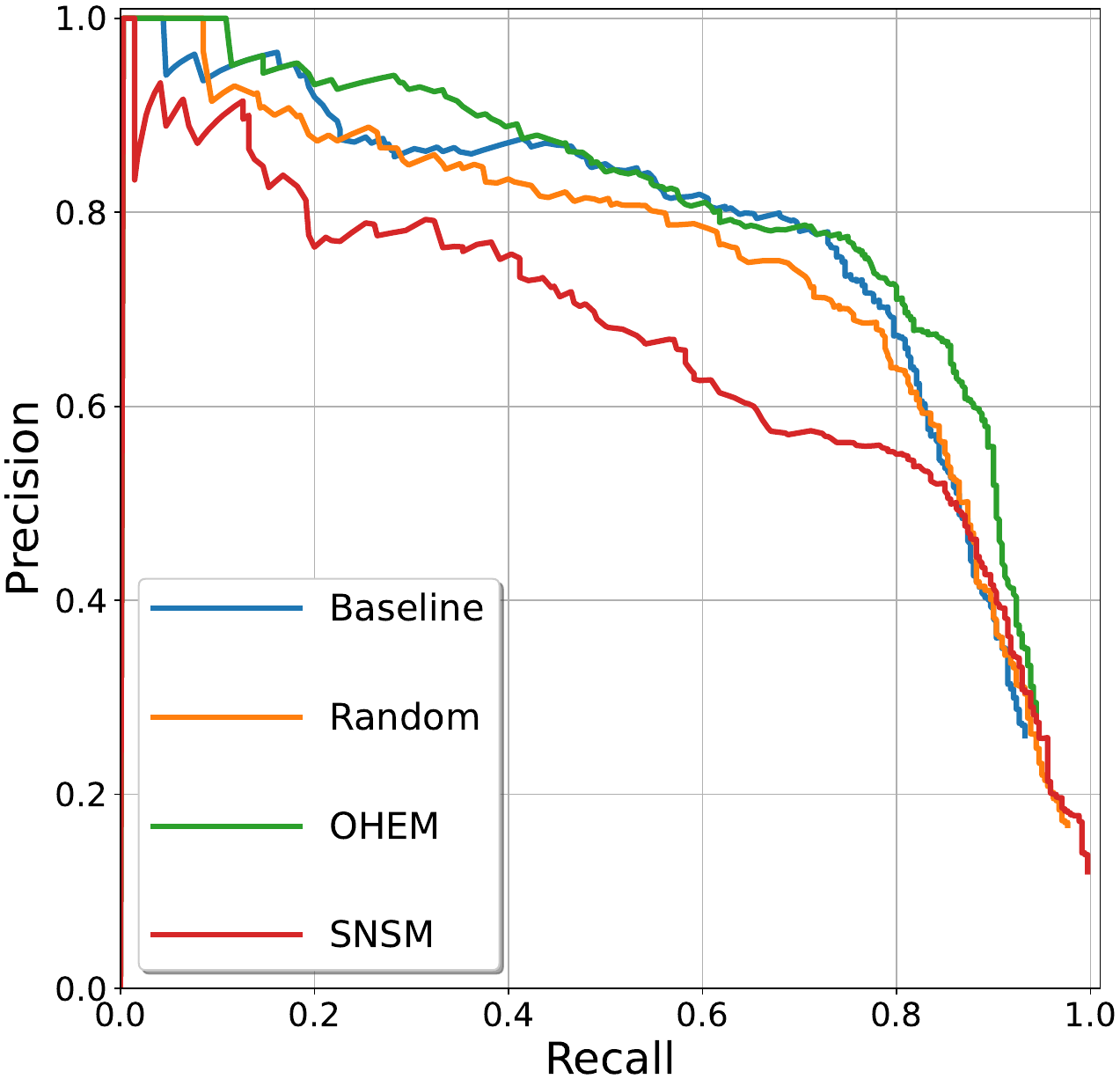}\label{SNSMfigure_4}}\quad
		\subfloat[][Video ROC Curve.]{\includegraphics[width=0.32\linewidth]{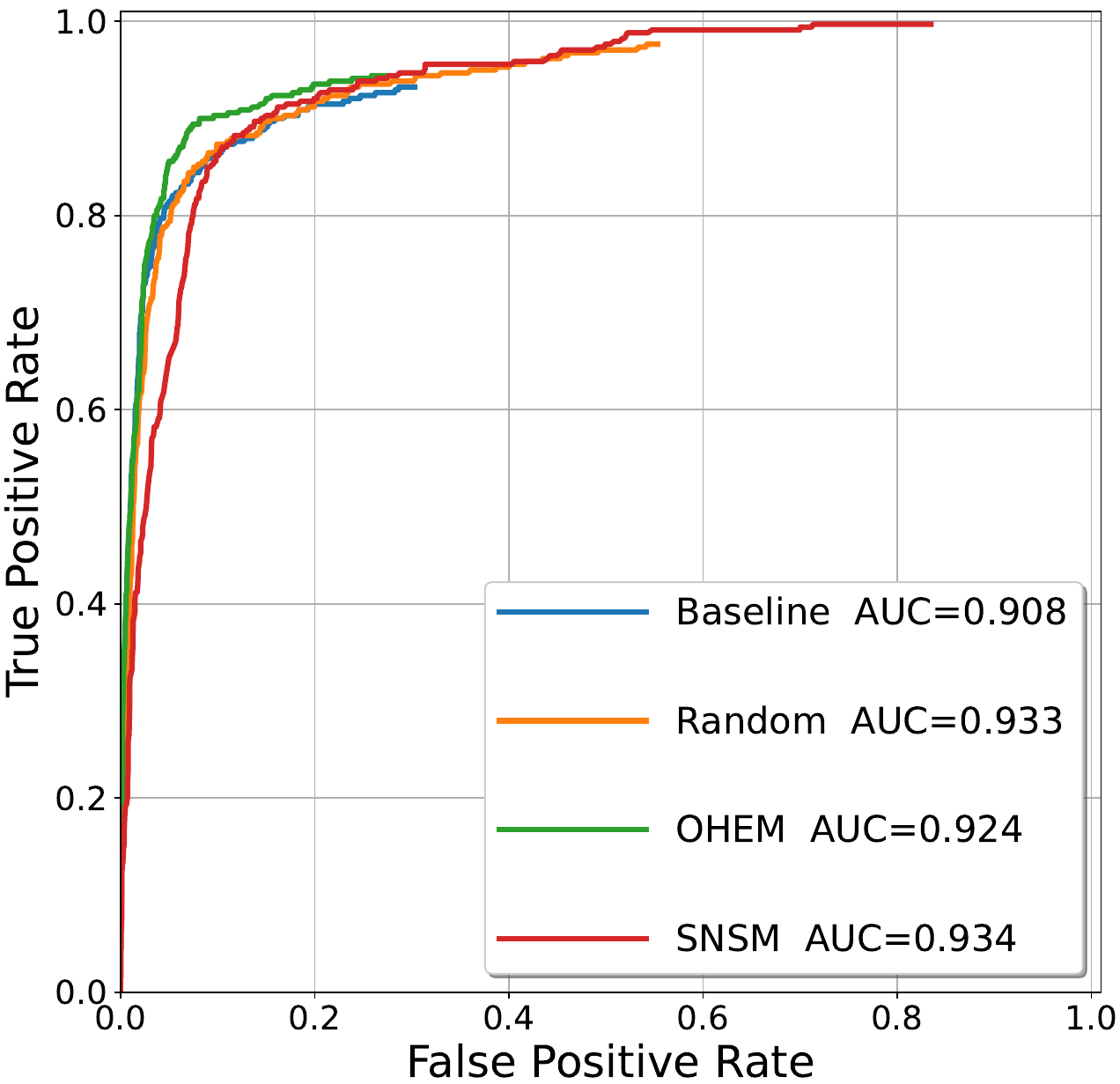}\label{SNSMfigure_5}}\quad		
           \subfloat[][Comparison of numerical metrics.]{\includegraphics[width=0.32\linewidth]{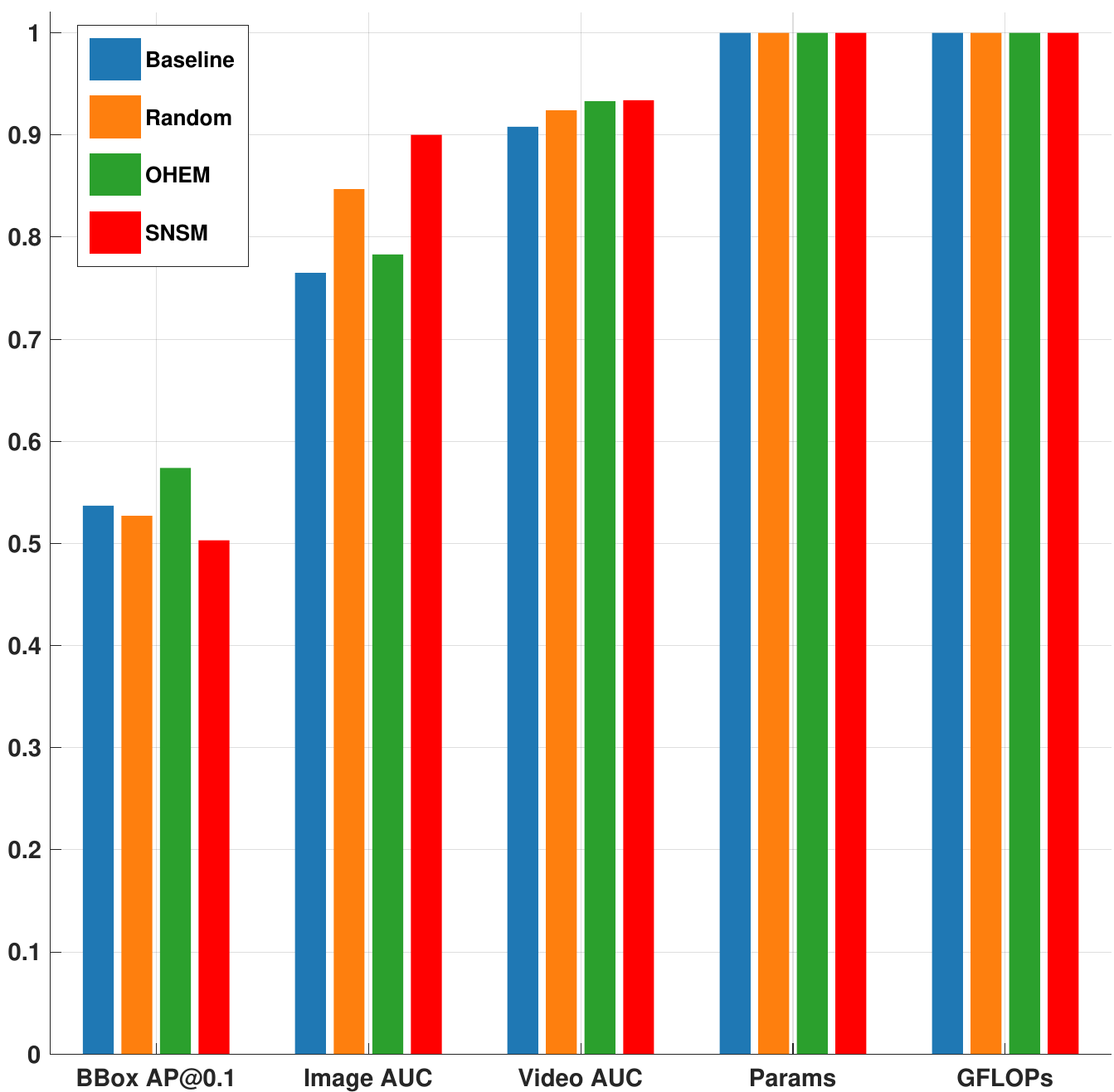}\label{hist_snsm}}\quad
	\end{minipage}
	\caption{Comparison of SNSM and other sampling mechanisms.(a),(b),(c),(d) and (e) are bounding box level PR Curve, image lever PR Curve, image level ROC Curve, video lever PR Curve, video level ROC Curve. (f) shows the numerical metrics comparison, where Params and GFLOPs are the normalized results.}
\label{fig:SNSM}
\end{figure*}

\paragraph{Effectiveness of Separable Negative Sampling.}
Separable negative sampling is designed for the problem of uncertain boundary of smoke instances. As shown in the Table~\ref{tab:SNSM}, the baseline model YOLOX-ContrastSwin does not employ any negative sampling strategy, that is, all locations participate in the training of the confidence branch. We experimented with randomly sampling 200 times the number of positive locations on negative locations, denoted as "Random". Similarly, we also tried to sample the negative locations with 200 times the number of positive locations using OHEM with the order of highest score to lowest score, denoted as "OHEM". Finally, SNSM is used to select the negative locations participating in the training, denoted as "SNSM". As can be seen from the results in the table, after adopting SNSM, the AP of bounding box decreases from 0.537 to 0.503, while the AUC values of image level and video level are absolutely increased by 13.5\% and 2.6\% respectively. The AP of bounding box level decreases because although SNSM can alleviate ambiguity and increase recall, it provides fewer negative training samples in positive images, which leads to inaccurate box positions. It can also be seen from the figure that SNSM does not perform well in the high-precision and low-recall interval, but it can make the model have relatively high precision in the high-recall interval. In practical fire applications, although false alarms degrade the user experience and are an industry pain point, ensuring high recall of fire alarms is still the first priority.

%\paragraph{Reproducibility Validation.} To validate the reproducibility of the model and the reliability of the results and conclusions, we conducted additional replicate experiments labeled as "run2" and "run3" in the last two rows of Table \ref{tab:ccpe} and Table \ref{tab:SNSM}. From the results in the tables, it can be observed that while multiple replicate experiments may yield fluctuating metric values, they do not alter the overall conclusions. The scale of our training and testing datasets is large enough to ensure such stability.

\subsection{Comparison to Classical Detectors}
It is a tradition to compare with classical methods. Since the code and datasets used in the researches of wildfire detection are rarely publicly available, we did not compare existing wildfire detection methods. Instead, we reproduce some classic and general object detectors such as YOLOV3~\cite{redmon2018yolov3}, YOLOX~\cite{ge2021yolox}, RetinaNet~\cite{lin2017focal}, Faster R-CNN~\cite{ren2016faster}, Sparse R-CNN~\cite{sun2021sparse}, on our own training and testing sets. As can be seen from the table below, the proposed model outperforms all existing models in terms of image and video level classification metrics without significant increase in the number of parameters and calculation. The metrics at the Bounding box level are slightly worse than the YOLOX model with CSPDarkNet backbone network in the 3rd and 4th row. There are two reasons for this phenomenon. First, the location and range of the smoke objects are controversial, so it is difficult to accurately detect. Secondly, SNSM reduces controversy by reducing the contribution of negative instances in positive image samples, which improves the recall rate of fire alarms, but also damages the fire clues localization inside positive image samples. However, we believe that in wildfire detection task, detecting fire event and raising alarm has the first priority, while locating the position of smoke or flame in the image has a lower priority although it is also important.

\begin{table*}
\begin{center}
\renewcommand\arraystretch{1.4}
\caption{Results of the proposed model and the other Classical Detectors on SKLFS-WildFire Test Dataset. The proposed model outperforms all existing models in both image and video level metrics. Bounding box level $AP@0.1$ are slightly worse than the YOLOX model with CSPDarkNet backbone network.}
\resizebox{0.99\textwidth}{!}{
\begin{tabular}{l|l|c|c|c||c|c}
\hline
Models\quad\quad\quad\quad\quad\quad \quad\quad\quad      &   Backbone\quad\quad\quad\quad\quad\quad      & \quad   BBox $AP@0.1$\quad    &  \quad  Image AUC \quad  & \quad  Video AUC\quad  & \quad  Params (M) \quad &\quad  GFLOPs\quad  \\
\hline\hline
YOLOV3     & DarkNet-53  &  0.421      &   0.588    & 0.823   &61.529  &139.972  \\
YOLOX    & CSPDarknet Small  & 0.504       &  0.679     &  0.882  &8.938  &26.640  \\
YOLOX         &   CSPDarknet Large & \textbf{0.506}       & 0.712      & 0.900   & 54.209 & 155.657 \\
YOLOX        &  Swin Tiny      & 0.476       &   0.732    &   0.909 & 35.840 & 50.524 \\
RetinaNet      & ResNet101        & 0.365       & 0.810      & 0.879   &56.961  &662.367  \\
Faster R-CNN     & ResNet101        & 0.432       & 0.598      &0.858    &60.345  & 566.591 \\
Sparse R-CNN    & ResNet101        & 0.505       & 0.852      &0.914    &125.163  &472.867  \\
OURS \quad\quad       & Swin Tiny      &  0.503     & \textbf{0.900}       &  \textbf{0.934}   & 35.893 & 53.250 \\
\hline
\end{tabular}
}
\label{tab:sota}
\end{center}
\end{table*}

\begin{figure*}[t]
	\centering
	\includegraphics[width=1\textwidth]{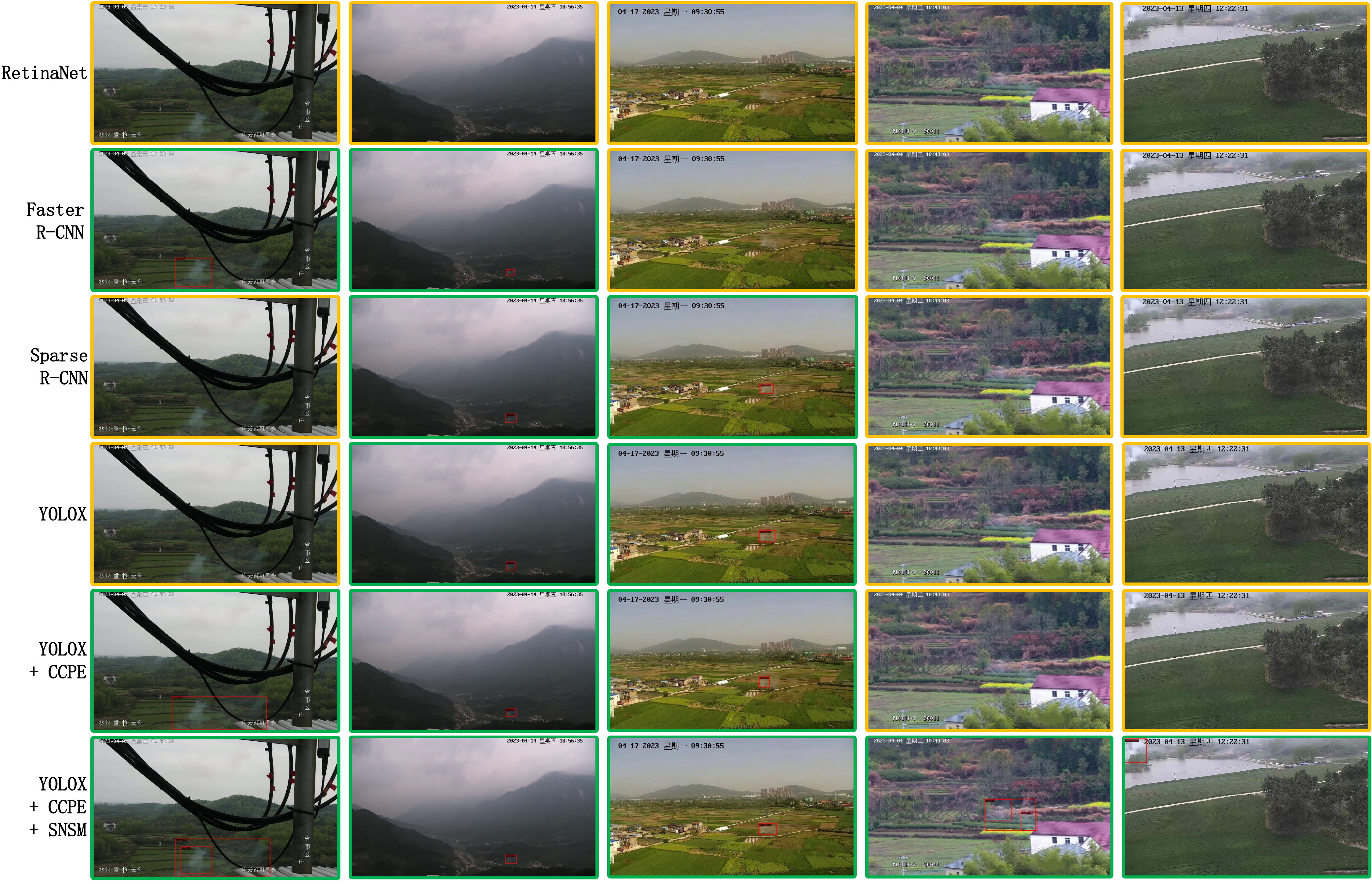}
	\caption{Detection results (score threshold is $0.5$) of the proposed model and baseline models on the SKLFS-WildFire Test Dataset. The green border indicates that the wildfire smoke objects are correctly detected, while the orange indicates that the wildfire smoke objects are missed, and the red bounding boxes in the images represent the smoke detection results.}	
	\label{fig:Vis_comp}
\end{figure*}

\begin{figure}[!thb]
	\centering
	\includegraphics[width=0.49\textwidth]{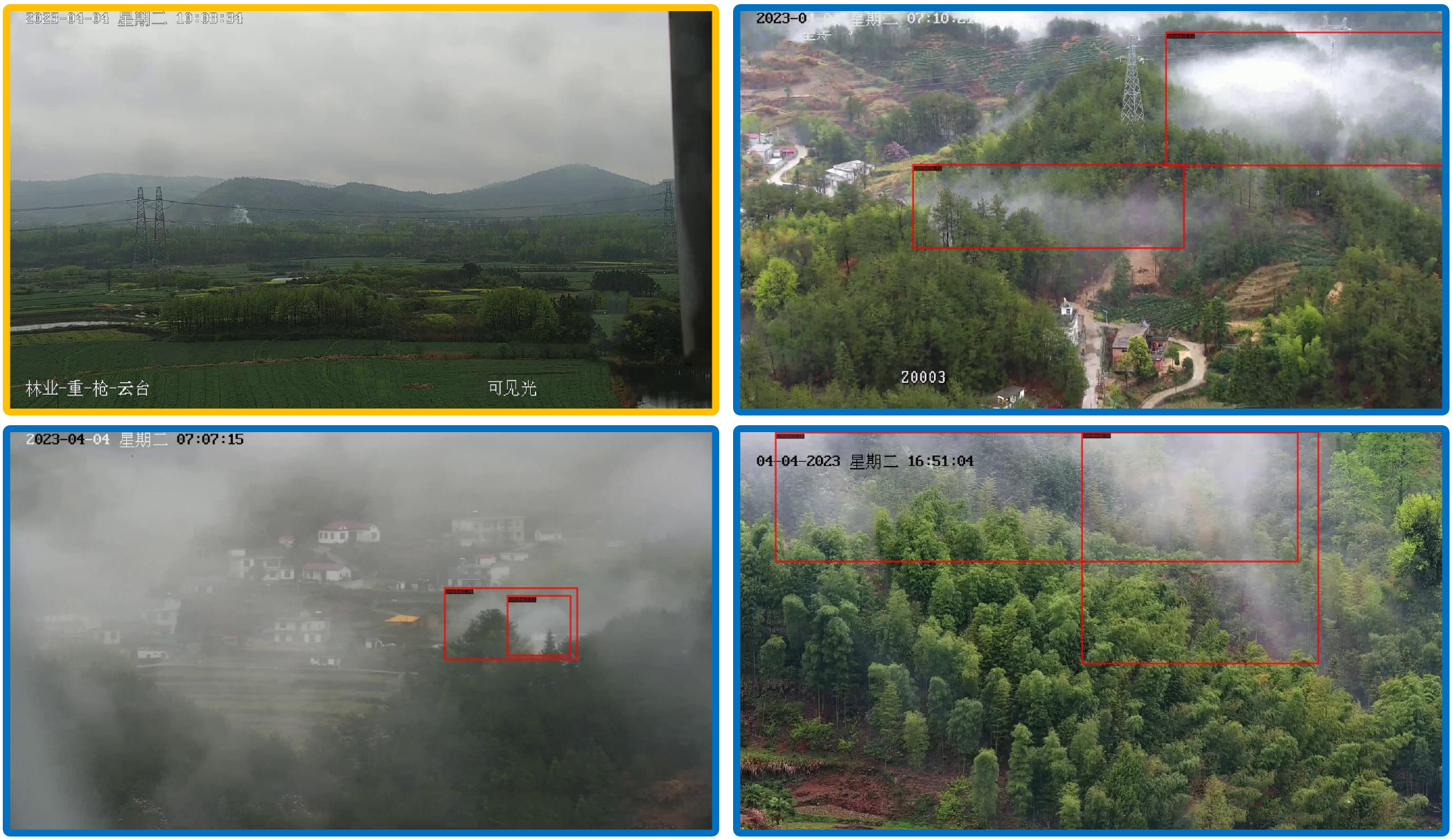}
	\caption{Bad cases of the proposed model. The orange indicates missing detection, while the blue indicates error detection. The complex background in open scenes is the cause of missed detection, while the clouds are the biggest interference factor that causes false detection.}	
	\label{fig:badcase}
\end{figure}

\subsection{Visualization Analysis}
Figure~\ref{fig:Vis_comp} shows the detection results of the proposed model and other baseline models on some samples of the SKLFS-WildFire Test Dataset. The score threshold is set to 0.5. The green image border indicates that the wildfire smoke objects are correctly detected, while the orange indicates that the wildfire smoke objects are missed, and the red bounding boxes in the images represent the smoke detection results. As can be seen from the figure, the proposed model has a huge advantage in terms of recall. The advantage of recall rate is precisely derived from the alleviation of the ambiguity problem of smoke box location and range by SNSM. However, as can be seen from the results of the last row, the high recall rate comes at the cost of multiple detection boxes being repeated in the smoke images, and these detection boxes are difficult to suppress with NMS. This phenomenon also confirms from the side that the problem of positional ambiguity of smoke objects does exist. The model generating multiple detection boxes for a single smoke instance will lead to a relatively low bounding box level metrics, such as AP@0.1, which is exactly the problem shown in Table~\ref{tab:SNSM} and Tabel~\ref{tab:sota}.

Figure \ref{fig:badcase} shows the error samples of the proposed model, where the yellow border indicates missed detection and blue border indicates false detection. The small number of missed detections may be due to the fact that the training data contains a lot of background interference similar to the appearance of smoke. It can be seen from the figure that under the interference of cloud and fog, it is easy to mistakenly identify the cloud and fog as smoke. In general, temporal features can distinguish interference with similar appearances, including clouds. Therefore, our next work is to use spatio-temporal information to achieve a more robust detection of wildfire smoke.

\section{Conclusion}
The performance of Transformer model in wildfire detection is not significantly better than that of CNN-based models, which deviates from the mainstream understanding of general computer vision tasks. Through analysis, we find that the advantage of the Transformer model is to capture the long-distance global context, and the ability to model the texture, color, and transparency of images is poor. Furthermore, the main clue of fire smoke discrimination lies in these low-level details. Therefore, this paper proposes a Cross Contrast Patch Embedding module to improve the Swin Transformer. Experiments show that the CCPE module can significantly improve the performance of Swin in smoke detection. 

Another main difference between smoke detection and general object detection is that the range of smoke is difficult to determine, and it is difficult to distinguish single or multiple instances of smoke. The assignment mechanism of negative instances in traditional object detectors can lead to contradictory supervision signals during training. Therefore, this paper proposes SNSM, which separates positive and negative image samples and uses different mechanisms to sample negative instances to participate in training. Experiments show that SNSM can effectively improve the recall rate of smoke, especially in the image level and video level metrics. However, the method proposed in this paper detects smoke based on static images, and through analysis, it is found that this method is difficult to eliminate the interference of cloud and fog between mountains. Building a fire detection model based on spatio-temporal information is the next step. 

Further, we relieased a new early wildfire dataset of real scenes, the SKLFS-WildFire Test, which can comprehensively evaluate the performance of wildfire detection model from three levels: bounding box, image, and video.  Its publication can provide a fair comparison benchmark for future research, whether it is detection or classification, static image schemes or video spatiotemporal schemes, and boost the development of the field of wildfire detection.

\section*{Acknowledgments}
This work was financially supported by the National Natural Science Foundation of China under Grant No. 32471866, the Anhui Provincial Science and Technology Major Project under Grant No. 202203a07020017, the Central Government-Guided Local Science and Technology Development Funds of Anhui Province under Grant No. 2023CSJGG1100 and the Key Project of Emergency Management Department of China under Grant No. 2024EMST010101. The numerical calculations in this paper have been done on the supercomputing system in the Supercomputing Center of University of Science and Technology of China. The authors gratefully acknowledge all of these supports.

\section*{Data statement}
The code and data can be directly accessed via \url{github.com/WCUSTC/CCPE}, or you can contact the corresponding author/first author for assistance. 
%

\printcredits

\bibliographystyle{cas-model2-names}
\bibliography{mybib}



\end{document}